\documentclass[10pt,twocolumn,letterpaper]{article}

\usepackage[table,xcdraw,dvipsnames]{xcolor}  
\usepackage{tabularx}
\usepackage{mdframed}
\usepackage{float}
\usepackage{titletoc}
\usepackage[pagenumbers]{iccv} 

\definecolor{iccvblue}{rgb}{0.21,0.49,0.74}
\usepackage[pagebackref,breaklinks,colorlinks,allcolors=iccvblue]{hyperref}

\def\paperID{6871} 
\def\confName{ICCV}
\def\confYear{2025}

\title{From Panels to Prose: Generating Literary Narratives from Comics}

\author{
  Ragav Sachdeva\quad Andrew Zisserman\vspace{5pt}\\
  {\normalsize Visual Geometry Group, Dept.\ of Engineering Science, University of Oxford}
}

\begin{document}

\twocolumn[{
    \vspace{-30pt}
    \renewcommand\twocolumn[1][]{#1}
    \maketitle
    \centering
    \vspace{-10pt}
    \includegraphics[width=\textwidth]{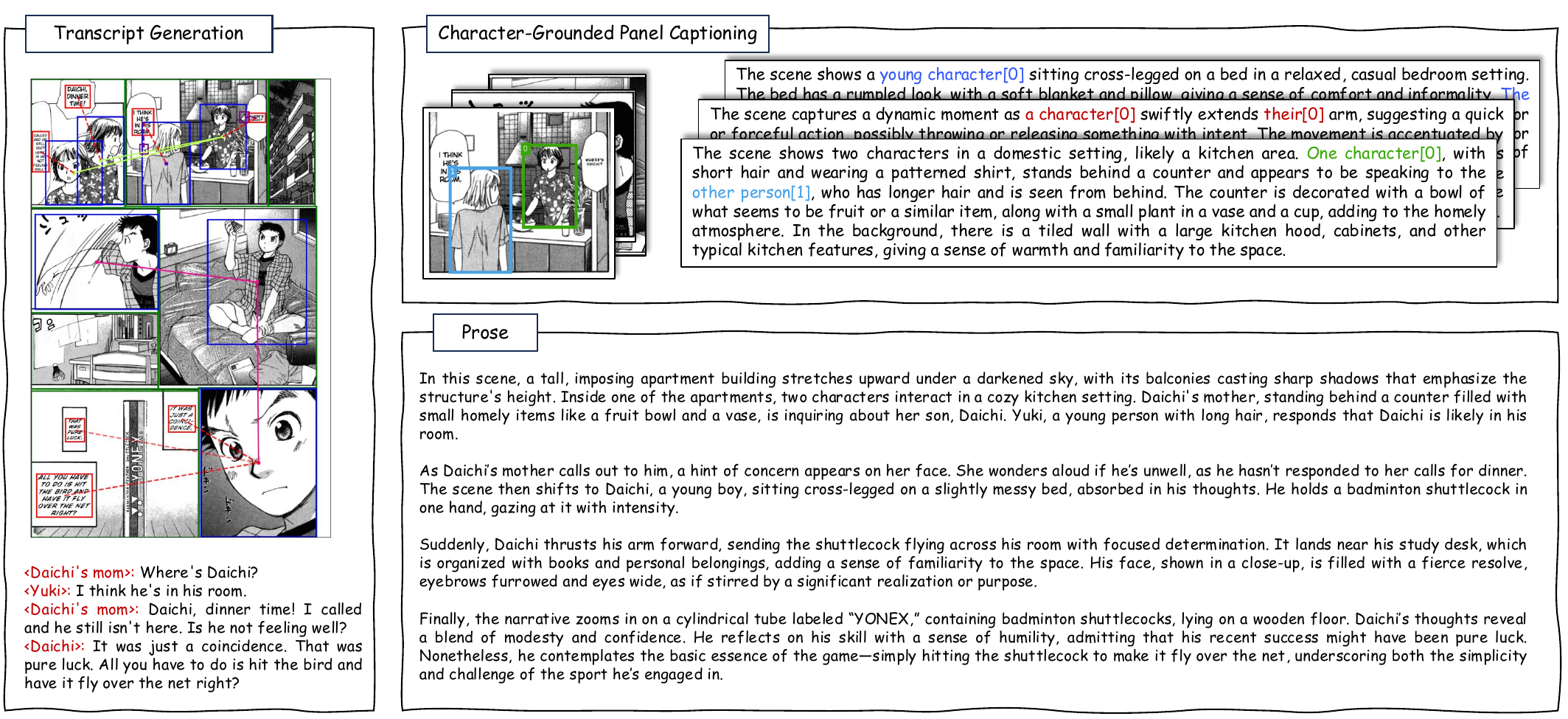}
    \captionof{figure}{
    Our approach to transforming comics into accessible narratives begins with generating transcripts (left), capturing dialogue. In the image, green boxes represent panels, blue boxes represent characters, red boxes represent text, and purple boxes represent speech-bubble tails. Solid lines indicate character clusters, while dashed lines show associations between dialogues and their speakers. This is followed by character-grounded panel captioning (top-right), where grounded phrases are colour-coded, and their corresponding predicted character boxes are overlaid on the panel images, adding descriptions and placing characters in context. Finally, these elements are combined into prose (bottom-right), creating a rich, immersive narrative for visually impaired readers.
    Images ©YamatoNoHane by Saki Kaori.
    }
    \label{fig:teaser}
    \vspace{20pt}
    }
    ]

\begin{abstract}
Comics have long been a popular form of storytelling, offering visually engaging narratives that captivate audiences worldwide. However, the visual nature of comics presents a significant barrier for visually impaired readers, limiting their access to these engaging stories. In this work, we provide a pragmatic solution to this accessibility challenge by developing an automated system that generates text-based literary narratives from manga comics. Our approach aims to create an evocative and immersive prose that not only conveys the original narrative but also captures the depth and complexity of characters, their interactions, and the vivid settings in which they reside.

To this end we make the following contributions: (1) We present a unified model, Magiv3, that excels at various functional tasks pertaining to comic understanding, such as localising panels, characters, texts, and speech-bubble tails, performing OCR, grounding characters etc. (2) We release human-annotated captions for over 3300 Japanese comic panels, along with character grounding annotations, and benchmark large vision-language models in their ability to understand comic images. (3) Finally, we demonstrate how integrating large vision-language models with Magiv3, can generate seamless literary narratives that allows visually impaired audiences to engage with the depth and richness of comic storytelling. Our code, trained model and dataset annotations can be found at: \url{https://github.com/ragavsachdeva/magi}.

\end{abstract}

\section{Introduction}
\label{sec:intro}
Comics, as a unique blend of visual art and narrative, have captivated audiences for decades, transcending cultural and linguistic boundaries. This dynamic medium not only entertains but also communicates complex themes and emotions through a combination of imagery and text. However, despite their widespread popularity, comics pose significant accessibility challenges for visually impaired readers. Traditional comic formats, reliant on intricate illustrations and visual storytelling techniques, often alienate those unable to perceive the visual elements.

Recent works have advanced manga accessibility by generating dialogue transcripts from manga, extracting text from speech bubbles and linking it to characters to provide a basic storyline~\cite{magiv1, magiv2}. While valuable, this approach captures only part of the narrative, missing scene descriptions, character actions, and emotional cues essential to the story. A recent survey also shows that visually impaired readers desire descriptions of scenes, emotions, and character interactions, highlighting the need for a comprehensive approach~\cite{accesscomics}.
Our work addresses this gap by producing a richer representation of the manga narrative, converting each page's visual context into a descriptive, literary prose.

The task of manga-to-prose conversion is complex, requiring an understanding of manga’s unique visual storytelling, human behavior, character relationships, and narrative pacing. Unlike video, which flows through continuous motion, manga tells its story in static frames that capture key moments. Translating these frames into prose demands a grasp of language, character dynamics, and story flow, as well as the skill to connect and interpret isolated elements. This paper aims to develop an automated pipeline to capture these nuances and generate coherent prose, making manga narratives accessible to those who cannot perceive the visual content.

Our solution transforms manga pages into accessible prose through a structured, multi-step pipeline. First, we identify and associate essential elements on each page—panels, characters, text, and speech-bubble tails—linking characters with their dialogues and clustering recurring characters to ensure consistency. Optical Character Recognition (OCR) is then applied to extract text and generate a dialogue transcript in the original manga-reading order, maintaining coherence, similarly to prior works~\cite{magiv1, magiv2}. After this, each panel is captioned individually, capturing scenes, actions, and emotional cues. Importantly, characters mentioned in these captions are localised precisely within panels, grounding descriptions to specific visual regions and ensuring continuity across the panels.
To transform these structured captions and dialogue into a cohesive text narrative, we leverage zero-shot prompting of foundational language models. This approach enables the model to generate prose without requiring extensive pre-training on manga-specific data. Additionally, while we default to prose for its immersive qualities, our approach is flexible: the model can also produce alternative formats, such as screenplay-style scripts, allowing users to select a narrative style that best suits their preferences.

In summary, our contributions are as follows:
\begin{itemize}
\item We introduce Magiv3, a unified model for manga analysis that performs tasks such as (i) detecting panels, characters, texts, and speech-bubble tails, (ii) associating texts with their speakers and tails, and characters with other characters (i.e., clustering), (iii) performing OCR while implicitly providing the reading order, and (iv) grounding characters in captions to visual regions in the image—all within a single framework, providing state-of-the-art structured input for prose generation.
\item We propose a comprehensive manga-to-prose pipeline that transforms manga pages into coherent literary narratives, enhancing accessibility for visually impaired audiences. This approach leverages Magiv3 to understand key manga elements and employs zero-shot prompting with off-the-shelf VLMs/LLMs (both open-source and proprietary) to generate accurate and immersive prose.
\item We introduce PopCaptions\footnote{\textbf{A note on copyright.}  All images included in this manuscript are from Manga109~\cite{manga109} where explicit permission is granted by the authors for use in publications. PopCaptions dataset only provides annotations for images in existing public datasets. We discuss this further in the supplementary.},
a novel character-grounded captioning dataset specifically curated to evaluate the capabilities of vision-language models (VLMs) in manga captioning and prose generation. Additionally, we propose an LLM-based evaluation setup to systematically assess these capabilities.
\end{itemize}

\begin{figure*}[ht]
    \centering
    \includegraphics[width=\linewidth]{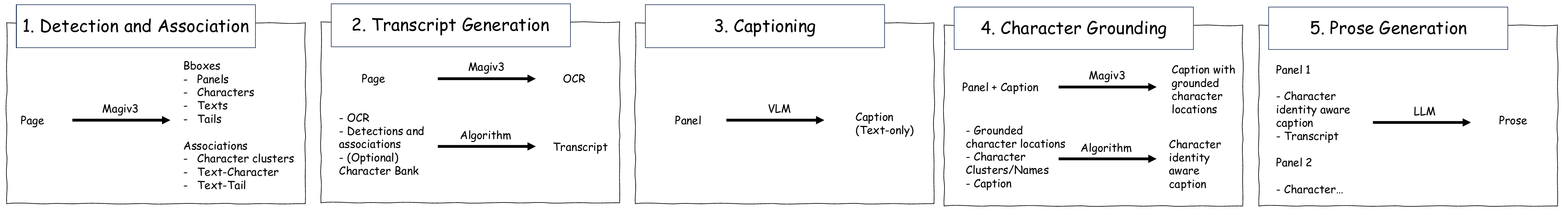}
    \caption{\textbf{Overview of the `Page to Prose' Pipeline.} The stages of the pipeline are described in section~\ref{sec:pipeline}. Magiv3 is described in section~\ref{sec:magiv3}, and the Captioning and Prose Generation in section~\ref{sec:text}.}
    \label{fig:pipeline}
\end{figure*}

\section{Related Works}
Comic understanding research spans tasks like panel detection~\cite{pang2014robust, ogawa2018object, he2018end, wang2015comic, nguyen2019comic, rigaud2015speech, magiv1, magiv2}, text and speech balloon detection~\cite{nguyen2019comic, piriyothinkul2019detecting, ogawa2018object, manga109coo, magiv1, magiv2}, character detection~\cite{topal2022domain, ogawa2018object, inoue2018cross, jiang2022decoupled, magiv1, magiv2}, re-identification~\cite{tsubota2018adaptation, qin2019progressive, zhang2022unsupervised, soykan2023identity, magiv2, li2024zeroshot}, and speaker identification~\cite{rigaud2015speech, manga109dialog, magiv1, magiv2, li2024zeroshot}. Key datasets, including Manga109~\cite{manga109, manga109coo, manga109dialog}, DCM~\cite{nguyen2018digital}, eBDtheque~\cite{guerin2013ebdtheque}, and PopManga~\cite{magiv1, magiv2}, have driven advances in these tasks, though methods traditionally used separate models for each, e.g., detection, OCR, and panel ordering~\cite{magiv1, magiv2}.

Recent efforts have aimed to standardise resources in comic analysis. The Comics Datasets Framework~\cite{vivoli2024comics} introduces Comics100, which broadens styles by including American comics and unified annotations for detection benchmarking. Expanding on this, CoMix~\cite{vivoli2024comix} provides a multi-task benchmark covering tasks like detection, re-identification, reading order, and multimodal reasoning, supporting model generalisation across diverse comic styles. Recent studies also leverage multimodal models to enhance contextual understanding~\cite{ikuta2024mangaub, vivoli2024comicap, soykan2024comicbert}.\newline

\noindent \textbf{Comparison with prior methods.} Our approach builds on the aforementioned advances, aiming to unify comic analysis tasks within a single framework to reduce reliance on multiple specialised models. We benchmark our method against publicly available datasets and methods in~\cref{sec:results} and further juxtapose the capabilities of our model with prior methods in the supplementary,

\section{Page to Prose Pipeline}
\label{sec:pipeline}
Given a collection of manga pages, our goal is to generate coherent literary prose that faithfully captures the original narrative and delivers an evocative reading experience. Manga images differ significantly from natural images due to their stylised, high-contrast visuals, non-standard layouts, and right-to-left reading orientation, which creates a unique distribution shift that conventional vision-language models struggle to interpret accurately. Furthermore, manga storytelling relies heavily on character continuity, emotional expression, and sequential panel-based narration, making the task of transforming static images into prose inherently complex. To address these challenges, we propose the multi-step pipeline shown in~\cref{fig:pipeline}, as described below.

\noindent\textit{\textbf{(1) Detection and Association.}} The initial step involves processing a manga page to localise the panels, characters, texts, and speech-bubble tails, and establish their relationships---character-character associations (i.e.\ clustering), text-character associations (i.e.\ speaker diarisation), and text-tail association (to assist with speaker identification).

\noindent\textit{\textbf{(2) Transcript Generation.}} Next, the dialogues are extracted using Optical Character Recognition (OCR), and along with the predicted speaker associations, a dialogue transcript is generated in the manga-reading order. This transcript may either include a unique ID for each character cluster, following the approach in Magi~\cite{magiv1}, or may include character names if a character bank is present, following the approach in Magiv2~\cite{magiv2}.

\noindent\textit{\textbf{(3) Caption Generation.}} Subsequently, each predicted panel is cropped and processed to generate a self-contained but informative caption, describing the scene, the characters, their actions, etc. This step is performed independently per panel, where each caption is generated by a general-purpose model using a short, concise zero-shot prompt that describes the image.

\noindent\textit{\textbf{(4) Character Grounding.}} Thereafter, each character referenced in the caption, e.g., ``... a character with black spiky hair ...'', is then localised in the panel image. This step is critical in tying the isolated panel descriptions to the broader storyline and linking the referenced characters to the identified clusters or named identities. This is accomplished by matching predicted boxes for grounded characters to the original boxes detected in step 1 using Intersection over Union (IoU) and bipartite matching, thus linking the grounded characters to their global identities.

\noindent\textit{\textbf{(5) Prose Generation.}} Finally, all of the information above is processed by an LLM to generate a prose. This step is essential for transforming the fragmented data from previous steps into a single coherent and readable narrative, allowing for a clearer understanding of the manga's storyline.

Despite the apparent complexity of solving numerous tasks outlined above, our solutions comprises of two models only:
(i) Magiv3, a unified model that can detect characters, texts, panels, speech-bubble tails, match dialogues to their speakers, cluster characters, ground characters in captions, perform OCR and implicitly order texts and panels, and (ii) a large foundational vision-language model which is instruction finetuned and used zero-shot for panel captioning and prose generation. More details are provided below.

\section{Detection, Association, OCR and Character Grounding}
\label{sec:magiv3}
\begin{figure*}[ht]
    \centering
    \includegraphics[width=0.99\linewidth]{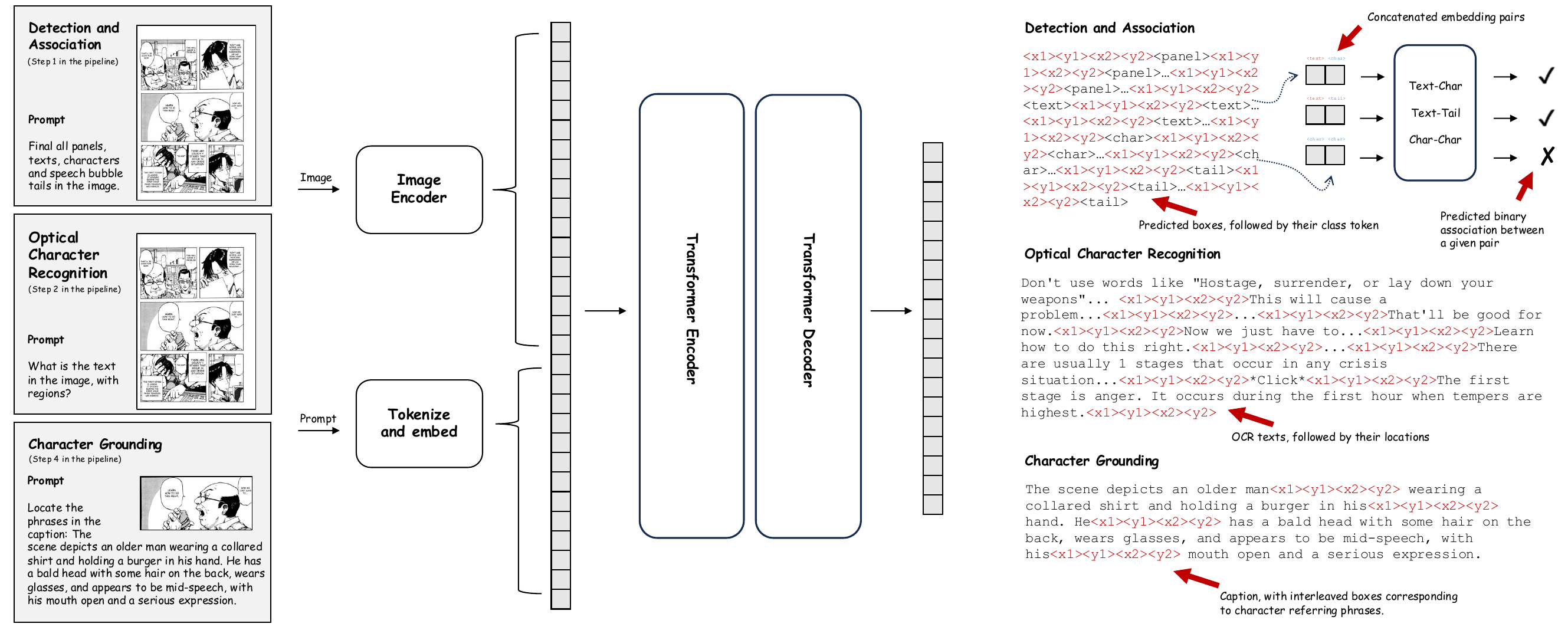}
    \caption{\textbf{The \textit{Magiv3} architecture and its three use cases.} The input to the model is an image and prompt pair. The output is text-only (tokens) predicted autoregressively for each of the three use cases. Images: ©HanzaiKousyouninMinegishiEitarou by Ki Takashi.}
    \label{fig:model_arch}
\end{figure*}

Previous approaches for manga analysis and transcript generation~\cite{magiv1,magiv2} rely on multiple specialised models: a DETR-based backbone for detection, a separate crop-embedding model to aid clustering, yet another model for OCR, and a post-processing algorithm for ordering elements through topological sorting. In contrast, we aim to achieve all these tasks (and more) within a single framework, using one set of model weights. Since the inputs and outputs can be heterogeneous depending on the task (e.g.\ image$\rightarrow$boxes for detection, image$+$text$\rightarrow$text$+$boxes for character grounding) we employ a transformer-based multi-modal encoder-decoder architecture and adopt a sequence-to-sequence framework, inspired by Florence-2~\cite{florence2}. In other words, given a high resolution manga image as input, along with a task-specific prompt, our model---\textit{Magiv3}---predicts the output as a sequence of tokens in an auto-regressive fashion.~\cref{fig:model_arch} shows an overview.

\subsection{Magiv3 Architecture}
\noindent\textbf{Image Encoder.} A high resolution manga image $I\in \mathbb{R}^{3\times H\times W}$ is processed by a dual attention vision transformer~\cite{davit} which extracts dense image features $g\in \mathbb{R}^{c\times v}$.

\noindent\textbf{Prompt Encoder.} A text-based task prompt $T$ is first tokenized and emedded using a word-embedding layer, resulting in features $h\in \mathbb{R}^{c\times t}$.

\noindent\textbf{Multi-Modal Transformer Encoder.} A standard transformer encoder jointly processes the concatenated vision and language embeddings, $[g\mathbin\Vert h] \in \mathbb{R}^{c\times (v+t)}$, to output contextualised features $j\in \mathbb{R}^{c\times (v+t)}$.

\noindent\textbf{Auto-regressive Transformer Decoder.} A standard transformer decoder, with casual self-attention, cross attends to the input features $j$, and predicts the output tokens auto-regressively.

\noindent\textbf{Association Heads.} Additionally, there are three association heads, which are simple MLPs similar to~\cite{magiv2}, and are used to predict text-character, character-character and text-tail associations, as shown in ~\cref{fig:model_arch}.

\subsection{Task Formulations}
\label{sec:task_formulation}

We tackle the manga image analysis problem by breaking it down into three distinct tasks: detection and association, OCR, and character grounding. Each of these tasks is handled independently, with a different prompt for each, but using the same set of model weights. This approach offers two main advantages: first, some tasks are inherently sequential (e.g., panel-level captions require predicting panel boxes first), and second, handling them separately enables more manageable context length for the model. These are the three tasks that Magiv3 is used for in the pipeline of the previous section, as illustrated in~\cref{fig:pipeline}.

\noindent\textbf{\textit{(1) Detection and Association:}} Given a manga page as input, the model is expected to output bounding boxes for panels, characters, texts, and speech-bubble tails, and to associate texts to characters, characters to other characters, and texts to tails. This is done with a single prompt, as shown in \cref{fig:model_arch}. Similar to prior works~\cite{magiv1,magiv2}, we formulate this as a graph generation problem, where the predicted boxes serve as the nodes, and the predicted associations form the edges. The model outputs the location tokens for bounding boxes, representing their quantized coordinates using 1000 bins (similar to~\cite{florence2, chen2021pix2seq, chen2022unified}), followed by their \textit{class} tokens (one of \texttt{<panel>}, \texttt{<char>}, \texttt{<tail>}, \texttt{<text>}). Since autoregressive generation inherently follows a specific order, we designed Magiv3 to output the boxes in manga-reading order (top to bottom, right to left), eliminating the need for an extra ordering step. These form the \textit{nodes} of the graph.

To obtain the \textit{edges}, we do not predict any edge-specific tokens, since the number of edges could grow quadratically in relation to the number of nodes. Instead, we take the feature embeddings corresponding to each predicted \textit{class} token from their last hidden state and process them in pairs via the association MLPs, resulting in a binary score that represents whether there is an edge between two nodes.

\noindent\textbf{\textit{(2) OCR:}} The second task involves extracting all the text from the manga page, along with its locations in the image, as shown in \cref{fig:model_arch}. This is done by running the model a second time with a different prompt. The model is tasked with outputting the text in the reading order, along with their corresponding bounding box coordinates. An example is shown in~\cref{fig:model_arch}. It is important to note that the text locations are output twice: once in the detection step (critical for associating texts with characters or speech bubbles), and once here in the OCR step, to associate the text with the corresponding location. The two outputs are reconciled easily by IoU based Hungarian matching.

\noindent\textbf{\textit{(3) Character Grounding:}} In this step, given a manga panel and its caption as input, the model is expected to ground all the characters referred to in the caption. This process involves identifying character-referencing phrases in an open-vocabulary setting, such as `boy', `girl', `mysterious figure', `character on the right', `bald man', as well as pronouns like `he', `she', and `they'. The model is tasked with locating these referenced characters within the panel, as shown in~\cref{fig:model_arch}. Please see the supplementary on an alternative way to solve this character grounding problem.

\subsection{Implementation}
The input to the model is an image $I$ and a text prompt $T$. The image $I$ is resized to $768\times768$px and processed by the DaViT~\cite{davit} model, resulting in flattenend ``visual token embeddings'' $g\in \mathbb{R}^{1024\times 577}$. The text prompt $T$ is tokenized and embedded using a word-embedding layer, resulting in ``prompt token embeddings'' $h\in \mathbb{R}^{1024\times t}$. The image token embeddings and the prompt token embeddings are concatenated resulting in the multi-modal input. This concatenated input is fed into a standard encoded-decoder transformer~\cite{vaswani2017attention}, comprising 12 layers each, with hidden dimension of 1024 and 16 attention heads. This transformer generates the output tokens in an auto-regressive fashion. In the case of edge prediction, the final features corresponding to \textit{class} tokens are processed in pairs by 3-layered MLPs to make a binary prediction towards the presence or absence of the edge. The model was initialised used Florence-2 weights~\cite{florence2} and trained on $2\times$ A40 GPUs, with an effective batch size of 32, using AdamW~\cite{adamw} optimiser with learning rate of 1e-4.

\section{Captioning and Prose Generation}
\label{sec:text}
Complementing the structural and relational analysis, the other part of our pipeline focuses on the generation of descriptive and narrative text, transforming the manga's visual elements into a flowing literary form. While manga images are rich in expressive details and complex character interactions, faithfully capturing this information in text is challenging, especially due to the lack of suitable training data. Therefore, we leverage existing large vision-language models (\textit{both open-source and proprietary}) to perform these tasks in a zero-shot setting.

\subsection{Caption Generation}
\label{sec:captioning}
\begin{figure*}[ht]
    \centering
    \includegraphics[width=\linewidth]{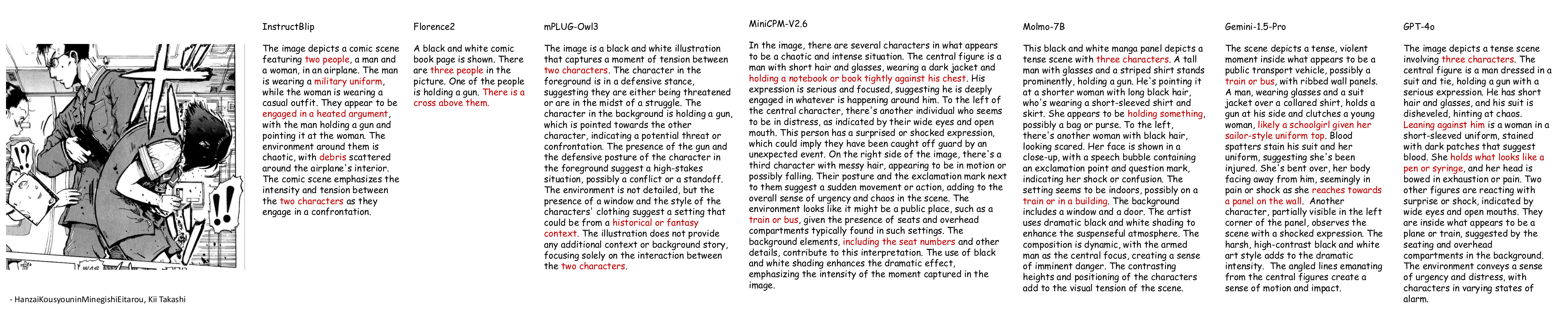}
    \caption{\textbf{Caption Comparison.}  We show the captions predicted by various vision-language models, both open-source and proprietary,  on a manga panel. The mistakes are highlighted in red. Overall, all the models make mistakes to some degree, such as miscounting the number of characters (there are four; notice the face below the `!?' on the left), hallucinating objects e.g.\ ``cross'', ``notebook'' etc.\ or incorrectly identifying the setting (which is an airplane). However, the general trend is that the captions get more accurate from left to right.
    }
    \label{fig:caption_comparison}
\end{figure*}

\begin{figure*}[ht]
    \centering
    \includegraphics[width=\linewidth]{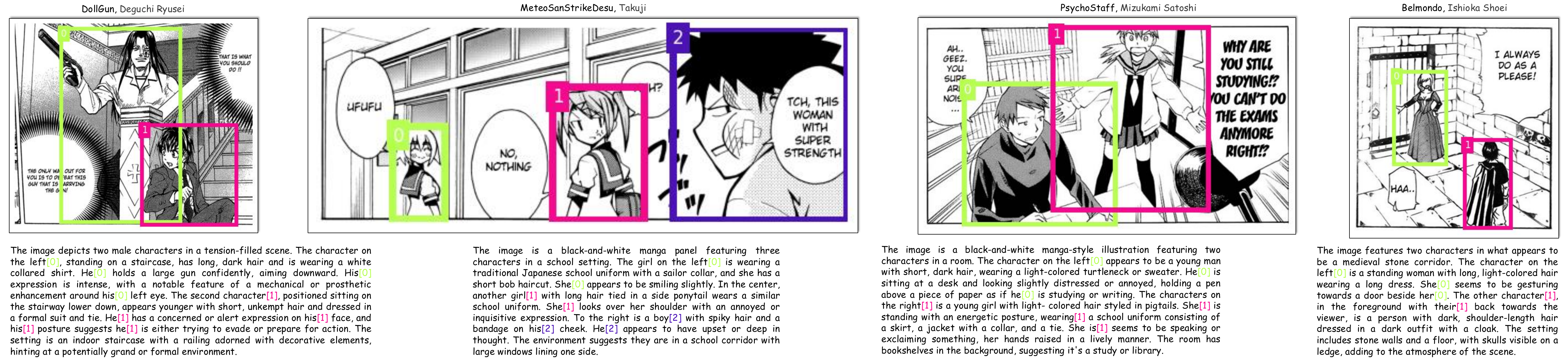}
    \caption{\textbf{Character-Grounded Panel Captions.}
    We show the captions predicted by GPT-4o-2024-08-06 on various manga panels and visualise the bounding boxes for characters grounded by our model (colour coded and numbered for visualisation only).}
    \label{fig:grounded_captions}
\end{figure*}

The task of captioning manga panels is particularly challenging due to the lack of suitable training datasets. As such, we rely on zero-shot captioning results from off-the-shelf vision-language models. In the caption generation stage, we adopt a panel-by-panel approach, where each manga panel is processed individually to generate its caption. This method allows us to preserve high-resolution details and ensures the model's attention remains focused on the unique elements of each panel—such as the setting, character emotions, and background details—while explicitly instructing it to ignore any embedded text, which it generally misinterprets and misattributes. We provide further discussion around this in~\cref{sec:results}, and show example predictions in~\cref{fig:caption_comparison,fig:grounded_captions}.

\subsection{Prose Generation}
In this final stage of our pipeline, we generate coherent narrative prose that accurately reflects the manga's storyline. Given the lack of suitable training data for this task, we rely on an instruction-finetuned large language model in a zero-shot setup, leveraging its ability to synthesise context across multiple components. For each panel on a manga page, we input both the dialogue transcript—complete with speaker associations—and the generated caption, which provide the spoken content and visual details. These panel-specific inputs are organised in the correct reading order and fed as structured input to the model. The model is then tasked with creating a continuous narrative that flows across the panels, transitioning smoothly from one scene to the next. Since the transcripts and captions are ``character aware,'' with consistent labels (or names, if available) for characters across panels, the model ensures character continuity and coherence throughout the generated prose.

Beyond basic narrative cohesion, the prose generation process also allows for flexibility in style. The output can vary in length, tone, and level of complexity, ranging from concise, child-appropriate summaries to more elaborate or formal narratives. This flexibility is enabled by zero-shot prompting, allowing the model to adapt to different styles and requirements. As a result, the prose generation process can produce a virtually limitless range of outputs, tailored to different audiences or use cases. The exact prompt used to generate results in~\cref{fig:prose} is provided in the supplementary material.

\begin{figure*}[h]
    \centering
    \includegraphics[width=\linewidth]{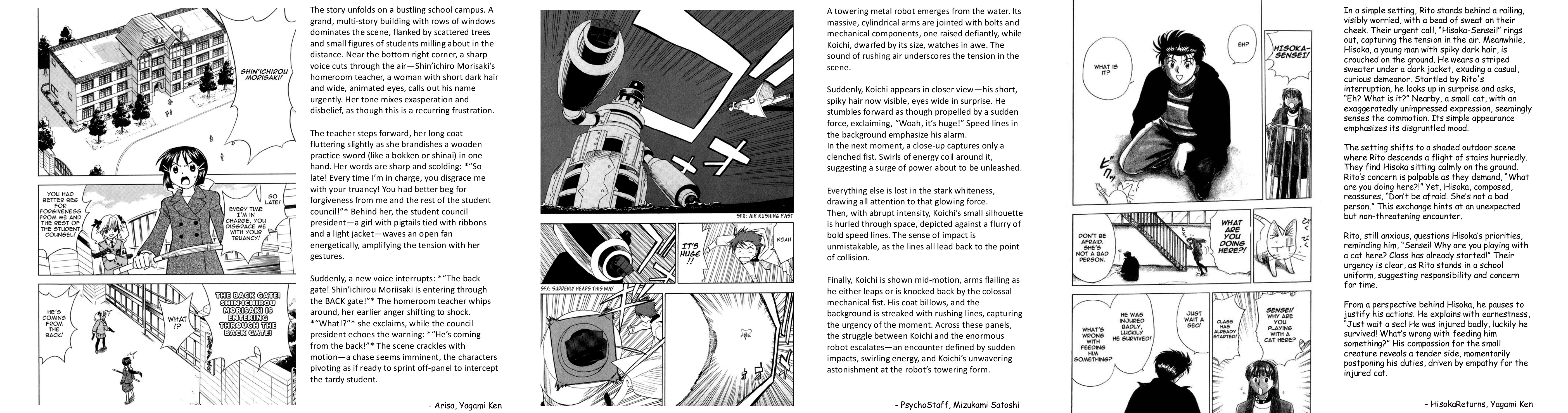}
    \caption{\textbf{Generated Prose Texts.} We show the final outcome of our pipeline, the generated prose, using various open-source and proprietary LLMs for Step 5 of the pipeline: Deepseek-R1 (left), o1-2024-12-17 (middle) and gpt-4o-2024-08-06 (right).}
    \label{fig:prose}
\end{figure*}

\section{PopCaptions: A Dataset for Character-Aware Manga Understanding}
\label{sec:popcaptions}

We introduce PopCaptions, a novel dataset of 3,374 captioned manga panels, drawn from 15 randomly selected chapters of the PopManga test set~\cite{magiv1} spanning 694 manga pages. These panels are diverse in content and style, and the labelled captions offer highly descriptive, character-aware annotations that go beyond conventional captioning datasets. Each caption is crafted to capture the unique storytelling style of manga, going beyond simple image descriptions like ``a black and white cartoon image'' to provide detailed accounts of character actions, emotions, and interactions within the scene. Additionally, our captions are specifically designed to link character mentions to precise locations within each panel, enabling advanced tasks in character grounding and visual narrative analysis. On average, there are 71.01 words per caption (median: 68, max: 222), with roughly 5.42 box instances in each of them (1.67 \textit{unique} instances). The vocabulary size (i.e.\ number of unique words) is 10644, while the total number of words across all captions is 239580.

In addition to the panel-level captions, we also provide page-level plot summaries (or prose) for each page of the 15 chapters. These character-centric caption and prose annotations makes PopCaptions a valuable resource for developing more nuanced and context-aware models that are sensitive to manga's unique visual and narrative structure. Please refer to the supplementary material for data annotation and quality assurance process.

\section{Results}
\label{sec:results}
We train our model, Magiv3, using Mangadex1.5M and the PopManga (dev) datasets~\cite{magiv1} for the detection, association, and OCR tasks. For character grounding, we automatically pseudo annotate some noisy training data using the approach outlined in~\cite{yang2023set}. We detail the collection, visualisations, and limitations of this data in the supplementary material. Captioning and prose generation are handled via off-the-shelf models, as elaborated further below.

We evaluate our system through both intermediate tasks and a single, composite measure of prose quality. This multi-step approach enhances transparency and interpretability: by decomposing the system into distinct sub-tasks (such as detection, captioning, and grounding), we can more easily identify the source of errors within the pipeline. For instance, if the generated prose is inconsistent, analyzing intermediate steps like clustering or grounding can help pinpoint underlying issues, which would be harder to isolate in a purely end-to-end evaluation. At the same time, assessing overall prose quality provides a holistic measure of the system's effectiveness. 

\subsection{Detection and Association}
In this section, we evaluate Magiv3's ability to detect key elements (panels, texts, characters, and tails) and correctly associate these elements through tasks such as text-to-speaker association, character clustering, and text-to-tail association. This evaluation is conducted on several benchmark datasets, including PopManga Test-S (PM-S), PopManga Test-U (PM-U), and Manga109 (M109). Quantitative results for detection performance are provided in~\cref{tab:detection}, and results for associations are shown in~\cref{tab:association}. Baseline comparisons are drawn against publicly available prior works to contextualise Magiv3's performance on these tasks.

The results in Tables \ref{tab:detection} and \ref{tab:association} demonstrate that Magiv3 outperforms previous models, across key tasks in manga image analysis. For detection, Magiv3 shows higher precision than prior works, while maintaining comparable recall, indicating improved accuracy in localising key elements within complex manga layouts. In association tasks, Magiv3 achieves the best scores for character clustering and text-to-tail associations, although Magiv2 slightly outperformed Magiv3 in speaker diarisation.

\subsection{Character Grounding}
In the character grounding stage, we evaluate the Magiv3 model's ability to accurately link character-referring phrases in captions (e.g.\ ``the character with dark hair'') to specific bounding boxes in the panel image that correspond to the intended character. In other words, the input to Magiv3 is a panel image and caption pair, and the output is the caption with interleaved boxes. Here, the goal is to localise uniquely identified characters based on descriptive clues in the caption, even in cases where multiple characters are present. The model must also carefully interleave bounding boxes within the caption text to prevent ambiguity when duplicate phrases refer to different characters across the panel. This is a more nuanced task than in prior grounding approaches~\cite{grounding_dino}, which would simply locate all characters in an image.

\noindent\textbf{Character grounding evaluation.} To evaluate character grounding, we first identify and match each character-referring phrase in the ground truth caption with its corresponding phrase in the model's predicted grounded caption. With these matches established, we assess the accuracy of the bounding boxes predicted for each phrase, effectively treating this as a detection task: determining whether the model accurately identified the location of each specified character. We report precision, recall and F1 scores on PopCaptions to capture the model's grounding accuracy. The results are shown in~\cref{tab:character_grounding} for both Magiv3, our primary model, and Magiv3-cg, a variant specifically finetuned for this task. As a baseline we use Florence-2's \textit{caption to phrase grounding} functionality, while acknowledging that it was not explicitly designed for this task and cannot differentiate between similar phrases referring to different characters. We discuss this further in the supplementary material.

\subsection{Captioning and Prose Generation}

\textbf{Captioning.} For captioning, we evaluate several vision-language models in a zero-shot setup, specifically: InstructBLIP~\cite{instructblip}, Florence-2~\cite{florence2}, mPLUG-Owl3~\cite{mplug_owl3}, MiniCPM-V2.6~\cite{minicpm}, Molmo-7B~\cite{molmo}, Gemini-1.5-Pro~\cite{gemini}, and GPT-4o~\cite{gpt4}. To ensure fairness, we maintain a consistent instruction prompt across all models (except InstructBlip\footnote{InstructBlip benefitted from a shorter, single line prompt (without the instruction to ignore all texts).} and Florence-2\footnote{Florence-2 uses its default prompt: ``Describe with a paragraph what is shown in the image.''}), which states: ``Describe this image to me. Focus on the characters, their appearance, their actions, and the environment. Please ignore any text, dialogues, or speech bubbles in the image.'' Note that this instruction format explicitly directs models to ignore text, as our initial observations indicated that these models tend to misinterpret or misattribute textual elements. Furthermore, dialogue information is already captured via Magiv3, making it unnecessary in captions, which should focus on the non-textual narrative context. Qualitative comparisons are given in~\cref{fig:caption_comparison,fig:prose}, while we detail the quantitative evaluation further below.

\noindent\textbf{Prose Generation.} This stage provides a holistic measure of our pipeline's effectiveness in generating coherent and immersive prose. Instead of evaluating subcomponents in isolation, we assess how well the structured information extracted by our system translates into high-quality narratives. For consistency, we use a fixed pipeline: Magiv3 handles detection, association, OCR, and character grounding, ensuring structured input. gpt-4o-2024-08-06 performs captioning, while DeepSeek-R1~\cite{deepseekr1}, a state-of-the-art LLM, synthesizes the final prose.

\begin{table}[H]
\resizebox{\columnwidth}{!}{%
\begin{tabular}{c|cc|cc|c|cc}
\hline
 &
  \multicolumn{2}{c|}{Characters} &
  \multicolumn{2}{c|}{Texts} &
  Panels &
  \multicolumn{2}{c}{Tails} \\ \hline
 &
  \multicolumn{1}{c|}{PM-S} &
  PM-U &
  \multicolumn{1}{c|}{PM-S} &
  PM-U &
  M109 &
  \multicolumn{1}{c|}{PM-S} &
  PM-U \\ \hline
DASS~\cite{topal2022domain} &
  \multicolumn{1}{c|}{\begin{tabular}[c]{@{}c@{}}0.80\\ \footnotesize\color[HTML]{9B9B9B}(0.92/0.70)\end{tabular}} &
  \begin{tabular}[c]{@{}c@{}}0.83\\ \footnotesize\color[HTML]{9B9B9B}(0.94/0.74)\end{tabular} &
  \multicolumn{1}{c|}{-} &
  - &
  - &
  \multicolumn{1}{c|}{-} &
  - \\
Florence2~\cite{florence2} &
  \multicolumn{1}{c|}{\begin{tabular}[c]{@{}c@{}}0.70\\ \footnotesize\color[HTML]{9B9B9B}(0.90/0.57)\end{tabular}} &
  \begin{tabular}[c]{@{}c@{}}0.73\\ \footnotesize\color[HTML]{9B9B9B}(0.91/0.60)\end{tabular} &
  \multicolumn{1}{c|}{\begin{tabular}[c]{@{}c@{}}0.62\\ \footnotesize\color[HTML]{9B9B9B}(0.93/0.46)\end{tabular}} &
  \begin{tabular}[c]{@{}c@{}}0.60\\ \footnotesize\color[HTML]{9B9B9B}(0.93/0.44)\end{tabular} &
  \begin{tabular}[c]{@{}c@{}}0.66\\ \footnotesize\color[HTML]{9B9B9B}(0.68/0.63)\end{tabular} &
  \multicolumn{1}{c|}{\begin{tabular}[c]{@{}c@{}}-\\ \footnotesize\color[HTML]{9B9B9B}(0/0)\end{tabular}} &
  \begin{tabular}[c]{@{}c@{}}-\\ \footnotesize\color[HTML]{9B9B9B}(0/0)\end{tabular} \\
Magi~\cite{magiv1} &
  \multicolumn{1}{c|}{\begin{tabular}[c]{@{}c@{}}0.80\\ \footnotesize\color[HTML]{9B9B9B}(0.80/0.81)\end{tabular}} &
  \begin{tabular}[c]{@{}c@{}}0.82\\ \footnotesize\color[HTML]{9B9B9B}(0.83/0.81)\end{tabular} &
  \multicolumn{1}{c|}{\begin{tabular}[c]{@{}c@{}}0.92 \\ \footnotesize\color[HTML]{9B9B9B}(0.95/0.89)\end{tabular}} &
  \begin{tabular}[c]{@{}c@{}}0.91\\ \footnotesize\color[HTML]{9B9B9B}(0.95/0.88)\end{tabular} &
  \begin{tabular}[c]{@{}c@{}}\textbf{0.93}\\ \footnotesize\color[HTML]{9B9B9B}(0.95/0.91)\end{tabular} &
  \multicolumn{1}{c|}{-} &
  - \\
Magiv2~\cite{magiv2} &
  \multicolumn{1}{c|}{\begin{tabular}[c]{@{}c@{}}0.81\\ \footnotesize\color[HTML]{9B9B9B}{\footnotesize\color[HTML]{9B9B9B}(0.83/0.80)}\end{tabular}} &
  \begin{tabular}[c]{@{}c@{}}0.83\\ \footnotesize\color[HTML]{9B9B9B}(0.86/0.81)\end{tabular} &
  \multicolumn{1}{c|}{\begin{tabular}[c]{@{}c@{}}\textbf{0.93}\\ \footnotesize\color[HTML]{9B9B9B}(0.96/0.90)\end{tabular}} &
  \begin{tabular}[c]{@{}c@{}}\textbf{0.92}\\ \footnotesize\color[HTML]{9B9B9B}(0.95/0.90)\end{tabular} &
  \begin{tabular}[c]{@{}c@{}}0.92\\ \footnotesize\color[HTML]{9B9B9B}(0.94/0.90)\end{tabular} &
  \multicolumn{1}{c|}{\begin{tabular}[c]{@{}c@{}}0.87\\ \footnotesize\color[HTML]{9B9B9B}(0.88/0.86)\end{tabular}} &
  \begin{tabular}[c]{@{}c@{}}0.86\\ \footnotesize\color[HTML]{9B9B9B}(0.88/0.85)\end{tabular} \\
Magiv3 [Ours] &
  \multicolumn{1}{c|}{\begin{tabular}[c]{@{}c@{}}\textbf{0.85}\\ \footnotesize\color[HTML]{9B9B9B}(0.92/0.79)\end{tabular}} &
  \begin{tabular}[c]{@{}c@{}}\textbf{0.87}\\ \footnotesize\color[HTML]{9B9B9B}(0.94/0.81)\end{tabular} &
  \multicolumn{1}{c|}{\begin{tabular}[c]{@{}c@{}}\textbf{0.93}\\ \footnotesize\color[HTML]{9B9B9B}(0.99/0.87)\end{tabular}} &
  \begin{tabular}[c]{@{}c@{}}\textbf{0.92}\\ \footnotesize\color[HTML]{9B9B9B}(0.98/0.87)\end{tabular} &
  \begin{tabular}[c]{@{}c@{}}0.92\\ \footnotesize\color[HTML]{9B9B9B}(0.95/0.90)\end{tabular} &
  \multicolumn{1}{c|}{\begin{tabular}[c]{@{}c@{}}\textbf{0.88}\\ \footnotesize\color[HTML]{9B9B9B}(0.89/0.88)\end{tabular}} &
  \begin{tabular}[c]{@{}c@{}}\textbf{0.90}\\ \footnotesize\color[HTML]{9B9B9B}(0.90/0.90)\end{tabular} \\ \hline
\end{tabular}%
}
\caption{\textbf{Detection results.} We report the F1-score, along with average precision and recall, of predicted boxes (post-threshold) on the PopManga (PM) S and U sets, and also on Manga109 (M109).
}
\label{tab:detection}
\end{table}

\begin{table}[H]
\resizebox{\columnwidth}{!}{%
\begin{tabular}{c|ccc|cc|cc}
\hline
       & \multicolumn{3}{c|}{Char-Char}                                                       & \multicolumn{2}{c|}{Text-Char}                         & \multicolumn{2}{c}{Text-Tail}                          \\ \hline
       & \multicolumn{1}{c|}{PM-S}            & \multicolumn{1}{c|}{PM-U}            & M109   & \multicolumn{1}{c|}{PM-S}            & PM-U            & \multicolumn{1}{c|}{PM-S}            & PM-U            \\ \hline
Florence2~\cite{florence2}   & \multicolumn{1}{c|}{-}          & \multicolumn{1}{c|}{-}          & - & \multicolumn{1}{c|}{-}          & -          & \multicolumn{1}{c|}{-}               & -               \\
Magi~\cite{magiv1}   & \multicolumn{1}{c|}{0.6574}          & \multicolumn{1}{c|}{0.6527}          & 0.6345 & \multicolumn{1}{c|}{0.5248}          & 0.5632          & \multicolumn{1}{c|}{-}               & -               \\
Magiv2~\cite{magiv2} & \multicolumn{1}{c|}{0.6745}          & \multicolumn{1}{c|}{0.6650}          & 0.6456 & \multicolumn{1}{c|}{\textbf{0.7499}} & \textbf{0.7512} & \multicolumn{1}{c|}{0.9838}          & 0.9830          \\
Magiv3 [Ours] & \multicolumn{1}{c|}{\textbf{0.6884}} & \multicolumn{1}{c|}{\textbf{0.6928}} &   \textbf{0.6782}     & \multicolumn{1}{c|}{0.7241}          & 0.7331          & \multicolumn{1}{c|}{\textbf{0.9911}} & \textbf{0.9859} \\ \hline
\end{tabular}
}
\caption{\textbf{Association results.} We report the AMI results for Char-Char associations and Average Precision as in~\cite{magiv2} for Text-Char and Text-Tail associations, on PM-S, PM-U and M109 datasets.}
\label{tab:association}
\end{table}

\begin{table}[H]
\centering
\scalebox{0.75}{
\begin{tabular}{c|c|c|c}
Captioner                  & F1 Score     & Precision     & Recall        \\ \hline
Florence2~\cite{florence2} & 0.28          & 0.25          & 0.32          \\
Magiv3 [Ours]              & 0.69          & 0.84          & 0.59          \\
Magiv3-cg [Ours]           & \textbf{0.74} & \textbf{0.89} & \textbf{0.63}
\end{tabular}%
}
\caption{\textbf{Character Grounding on PopCaptions.}
}
\label{tab:character_grounding}
\end{table}

\begin{table}[H]
\resizebox{\columnwidth}{!}{%
\begin{tabular}{c|cccc|c}
\hline
 &
  \multicolumn{4}{c|}{Judge} &
   \\ \hline
Captioner &
  \multicolumn{1}{c|}{GPT-4~\cite{gpt4}} &
  \multicolumn{1}{c|}{Gemini-1.5~\cite{gemini}} &
  \multicolumn{1}{c|}{Llama3~\cite{llama3}} &
  Gemma2~\cite{gemma2} &
  Avg \\ \hline
instructblip-vicuna-7b~\cite{instructblip} &
  \multicolumn{1}{c|}{1.97} &
  \multicolumn{1}{c|}{1.87} &
  \multicolumn{1}{c|}{2.45} &
  2.20 &
  2.12 \\
florence-2-large-ft~\cite{florence2} &
  \multicolumn{1}{c|}{2.15} &
  \multicolumn{1}{c|}{1.99} &
  \multicolumn{1}{c|}{2.27} &
  2.22 &
  2.16 \\
\multicolumn{1}{l|}{mPLUG-Owl3-7B-240728~\cite{mplug_owl3}} &
  \multicolumn{1}{c|}{2.60} &
  \multicolumn{1}{c|}{2.38} &
  \multicolumn{1}{c|}{3.39} &
  2.78 &
  2.79 \\
miniCPM-V-2\_6~\cite{minicpm} &
  \multicolumn{1}{c|}{2.86} &
  \multicolumn{1}{c|}{2.70} &
  \multicolumn{1}{c|}{3.47} &
  2.93 &
  2.99 \\
molmo-7B-D-0924~\cite{molmo} &
  \multicolumn{1}{c|}{3.13} &
  \multicolumn{1}{c|}{2.98} &
  \multicolumn{1}{c|}{3.77} &
  3.13 &
  3.25 \\
gemini-1.5-pro~\cite{gemini} &
  \multicolumn{1}{c|}{3.50} &
  \multicolumn{1}{c|}{\textbf{3.49}} &
  \multicolumn{1}{c|}{4.03} &
  3.48 &
  3.62 \\
gpt-4o-2024-08-06~\cite{gpt4} &
  \multicolumn{1}{c|}{\textbf{3.63}} &
  \multicolumn{1}{c|}{3.43} &
  \multicolumn{1}{c|}{\textbf{4.09}} &
  \textbf{3.49} &
  \textbf{3.66} \\ \hline\hline
\textit{Prose [Our pipeline]} &
  \multicolumn{1}{c|}{\textit{3.61}} &
  \multicolumn{1}{c|}{\textit{3.27}} &
  \multicolumn{1}{c|}{\textit{4.06}} &
  \textit{3.50} &
  \textit{3.61}
\end{tabular}%
}
\caption{\textbf{Captioning and Prose results on PopCaptions.} The scores are on a scale of 1-5, where 1 = Severely Inaccurate, 2 = Somewhat Off-Base, 3 = Partially Accurate, 4 = Mostly Accurate and 5 = Highly Accurate. The LLM judges used are two proprietary models (gpt-4o-2024-08-06 and gemini-1.5-pro), and two open-source models (llama3-70b-8192 and gemma2-9b-it).}
\label{tab:captioning}
\end{table}

\noindent\textbf{Evaluation.} Evaluating captions and prose poses unique challenges, as traditional metrics (e.g., CIDEr, METEOR) are not well-suited. Given the infinite ways to accurately describe a manga panel or narrate a story, rigid metric-based assessments can be misleading. Instead, we adopt a large language model (LLM) judge-based evaluation protocol~\cite{chan2023clair}, prompting LLMs to compare the predicted captions/prose with reference caption/prose and rate their similarity on a 1-5 scale, from \textit{Severely Incorrect (1)} to \textit{Highly Accurate (5)}. The prompt and rubric for this evaluation are provided in the supplementary material. Quantitative captioning and prose results are presented in~\cref{tab:captioning},  where we use multiple LLMs as independent judges.  Please see the supplementary material for some examples of LLM-based scoring.

The captioning results in Table \ref{tab:captioning} reveal notable disparities among vision-language models. GPT-4o and Gemini-1.5-Pro lead with average scores of 3.66 and 3.62, respectively, demonstrating strong alignment with reference captions and effective extraction of visual details. Open-source models, while still lagging in handling narrative-driven media like manga, show promising improvements—Molmo-7B, despite its smaller size, performs surprisingly well, highlighting steady advancements in open-source vision-language capabilities. 

For prose generation, our structured pipeline achieves an average score of 3.61—between 3 (Partially Correct) and 4 (Mostly Correct)—indicating that it produces largely accurate and coherent narratives. This quantification offers a concrete measure of performance, while upstream tasks—detection, association, captioning, and character grounding—serve as valuable indicators of prose quality by ensuring the structured representation of narrative elements. 

\subsection{Limitations}
While our system shows strong performance in manga analysis, character grounding still has substantial room for improvement. In complex scenes with numerous characters, the model's accuracy drops significantly, indicating a persistent challenge in maintaining high-quality, consistent grounding. This decline is primarily due to limitations in the training data, which suffers from both imbalance and annotation quality issues, particularly in scenes with overlapping characters. Addressing this in future work, with richer and more balanced datasets and improved annotation quality, could greatly enhance the model's grounding abilities and its utility for generating coherent prose narratives.

\section{Conclusion}
This work addresses the accessibility challenges of manga for visually impaired readers by developing a system that transforms manga into immersive, text-based narratives. We introduced Magiv3, a model capable of manga-specific tasks such as transcript generation and character grounding, enabling the generation of prose that captures the narrative richness of manga. Alongside this, we release a captioning and prose dataset with character-grounding annotations, providing a valuable resource for future comic understanding research.

\newpage
\noindent \textbf{Acknowledgements:} We would like to thank Cindy Seuk, Puja Sachdeva and Anhad Sachdeva for their assistance with caption quality assurance. This research is supported by EPSRC Programme Grant VisualAI EP/T028572/1 and a Royal Society Research Professorship RP\textbackslash R1\textbackslash191132.

{
    \small
    \bibliographystyle{ieeenat_fullname}
    \bibliography{main}
}

\clearpage
\onecolumn
\setcounter{page}{1}

\maketitlesupplementary

\appendix

\section{Distinctive contributions of the Magiv3 model}
While the main contribution of this work is the pipeline we propose as a whole, along with the dataset annotations which permit future research and benchmarking, in this section we want to draw attention to the proposed Magiv3 model. Since this model is built on top of Florence-2~\cite{florence2}, we want to be clear and transparent about how our model is distinctive from prior works.~\cref{tab:comparison} provides an overview.

\begin{table}[h]
\centering
\resizebox{\columnwidth}{!}{%
\begin{tabular}{c|cccccc}
 &
  Detection &
  Character Clustering &
  Text - Character Matching &
  OCR &
  Captioning &
  Character Grounding \\ \hline
DASS~\cite{topal2022domain}       & Characters                       & No  & No  & No  & No  & No         \\
Magiv1~\cite{magiv1} &
  Characters, Texts, Panels &
  Yes, using a separate character crop encoder &
  Yes &
  No, uses a separate model &
  No &
  No \\
Magiv2~\cite{magiv2} &
  Characters, Texts, Panels, Tails &
  Yes, using a separate character crop encoder &
  Yes &
  No, uses a separate model &
  No &
  No \\
Florence-2~\cite{florence2} & Zero-shot, performs poorly       & No  & No  & Yes, but line-by-line instead of text-blocks & Yes, low quality & Not really \\
Magiv3 [Ours]     & Characters, Texts, Panels, Tails & Yes & Yes & Yes & No  & Yes       
\end{tabular}%
}
\caption{\textbf{Comparison of Magiv3 with prior works.}}
\label{tab:comparison}
\end{table}

\begin{figure*}[h]
    \centering
    \includegraphics[width=\linewidth]{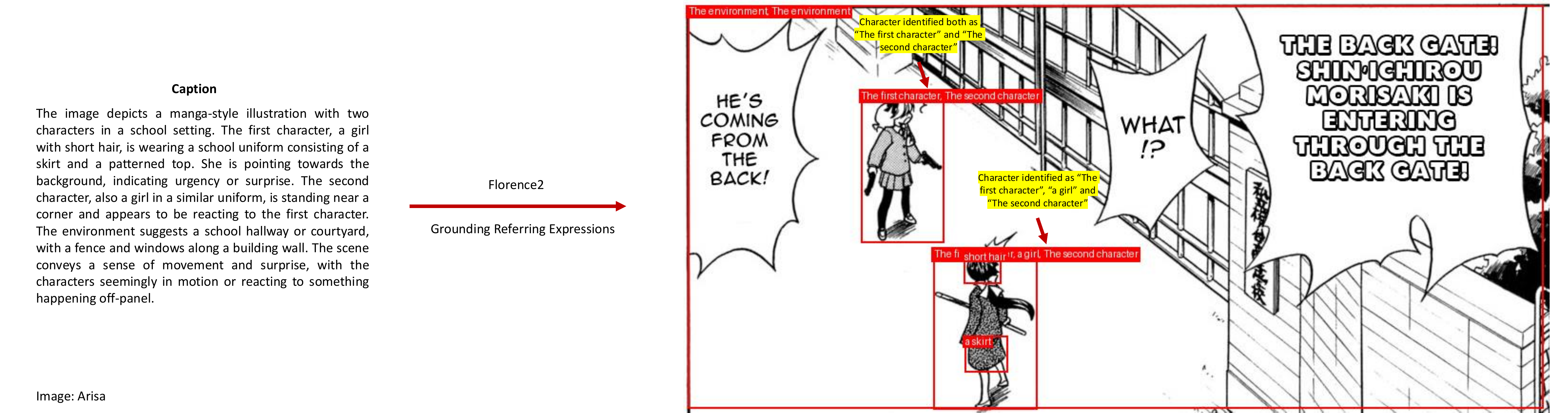}
    \caption{\textbf{Florence2 Character Grounding.} Florence2 is trained to perform the \textit{caption to phrase grounding task}. For this task, the input to the model is an image, along with its caption, and the model outputs a set of \texttt{[phrase, bbox]} pairs. We visualise these predicted phrases and the corresponding boxes (right). As noted in the text highlighted in yellow, Florence2 grounds the phrase ``The first character'' to two different characters, which is clearly a contradiction. This contradiction is also apparent for the phrase ``The second character''. This shows that Florence2's \textit{caption to phrase grounding} ability is only adept at extracting ``class labels'' from the captions (e.g. character) and predicting boxes for all locations where this class object appears. This behaviour is inadequate for the character grounding task proposed in this work where the model should disambiguate character referring phrases based on the context.}
    \label{fig:florence2_character_grounding}
\end{figure*}

The notable advantage of Magiv3 is that it is a unified architecture for many manga-related tasks with a single set of model weights. Despite being built on top of Florence-2, Magiv3's abilities are significantly different. For instance, as shown in~\cref{fig:florence2_character_grounding}, Florence-2 cannot meaningfully ground characters in the given captions. It has no notion of context specific phrase grounding and ultimately finds ``class labels'' from the captions and predicts boxes for them. Florence-2 also cannot perform character clustering or speaker diarisation which Magiv3 can. Conversely, Magiv3 has not been trained to generate captions due to the high training costs required, while Florence-2 can generate [low quality] captions.

\begin{figure*}[h]
    \centering
    \includegraphics[width=\linewidth]{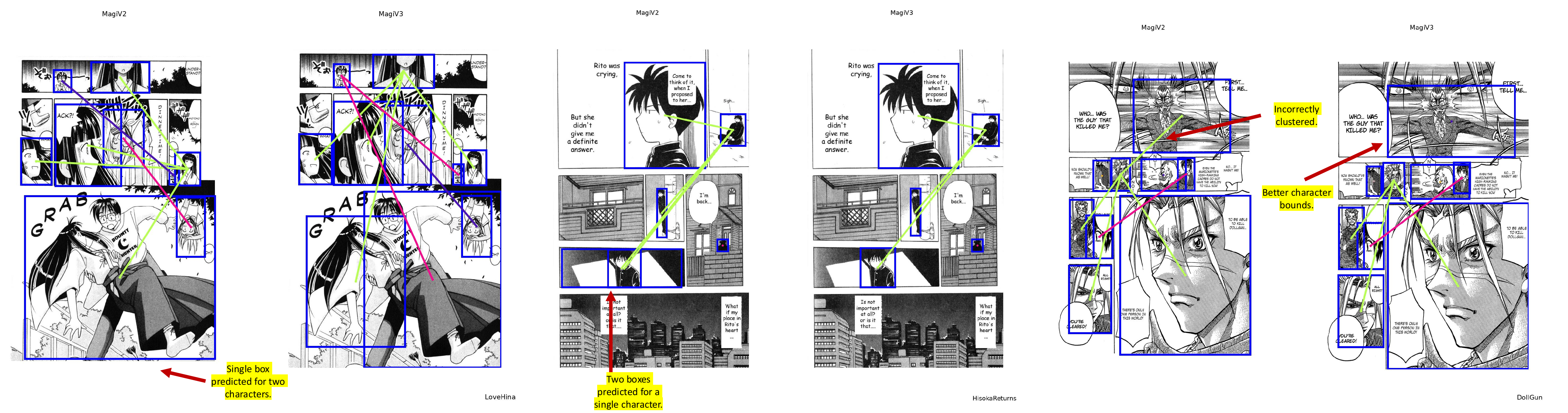}
    \caption{\textbf{Magiv2 vs Magiv3 character detection and clustering.} The predicted character boxes are highlighed in blue. The connected components formed by the colourful lines show the predicted character clusters. As noted in the text in yellow, Magiv2 can often miss predictions (grouping multiple characters in a single box) or over predict (predicting multiple boxes for a single character). This issue is less apparent in Magiv3, while also providing tighter boundaries for characters.}
    \label{fig:magiv2_vs_v3}
\end{figure*}

\begin{figure*}[h]
    \centering
    \includegraphics[width=\linewidth]{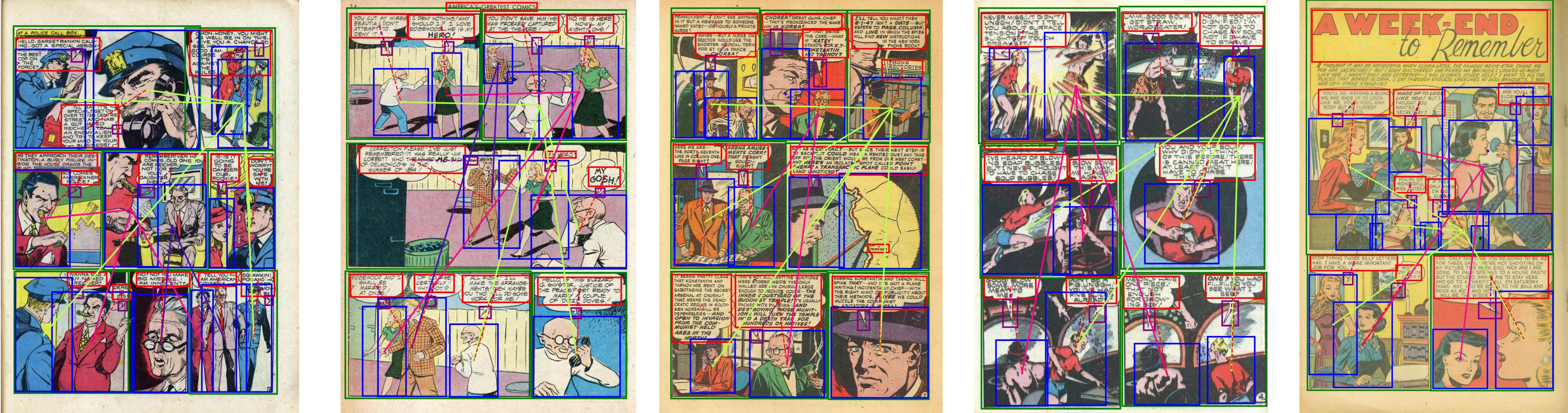}
    \caption{\textbf{Magiv3's predictions on Western Comics.} The predicted panels bboxes are in green, characters in blue, text boxes in red and speech bubble tails in purple. The character clusters are shown using lines connecting different characters (colour-coded to visualise connected components), the text-character associations are shown using dashed red lines and text-tail associations are shown using dashed purple lines. Images taken from~\cite{iyyer2017amazing}.}
    \label{fig:western_comics}
\end{figure*}

Additionally, Magiv3 can do everything Magiv1, Magiv2 etc. can (and more), without needing a separate crop encoder trained contrastively, or a separate OCR model, or post-processing for ordering. In fact, due to the next token prediction paradigm, we do not have to tweak any thresholds to filter out low confidence predictions; unlike prior works which use finicky hyperparameters which are sensitive to near-boundary predictions. As shown in~\cref{fig:magiv2_vs_v3}, the optimal threshold for character detection in Magiv2 tends to either miss predictions or add false positives.
In contrast, Magiv3 works much better and hence has a much higher precision for character detection (as noted in the main paper). Finally, even though we exclusively train our model on Japanese comics, in~\cref{fig:western_comics} we show that it works surprisingly well on Western comics as well.

\section{More on character grounding}

As mentioned in Sec. 4.2 of the main paper, here we explore an alternative, training-free, and zero-shot method for obtaining pre-grounded captions for manga panels. In fact this method is leveraged to acquire training data for Magiv3, as alluded to in Sec. 7 of the main paper. Finally, we discuss the inherent limitations of this technique.

The core idea behind this approach is simple yet effective. First, we detect all characters in a manga page using an existing model, thereby identifying the bounding boxes of all characters within individual panels. This ensures that the spatial locations of the characters are already well-defined. Subsequently, inspired by the visual prompting methodology in~\cite{yang2023set}, we overlay the detected bounding boxes on the panel image and assign each bounding box a unique identifier. The modified panel, now augmented with visual prompts and unique IDs, is fed into GPT-4o-2024-08-06 alongside a carefully crafted instruction prompt. 

\begin{figure*}[h]
    \centering
    \includegraphics[width=\linewidth]{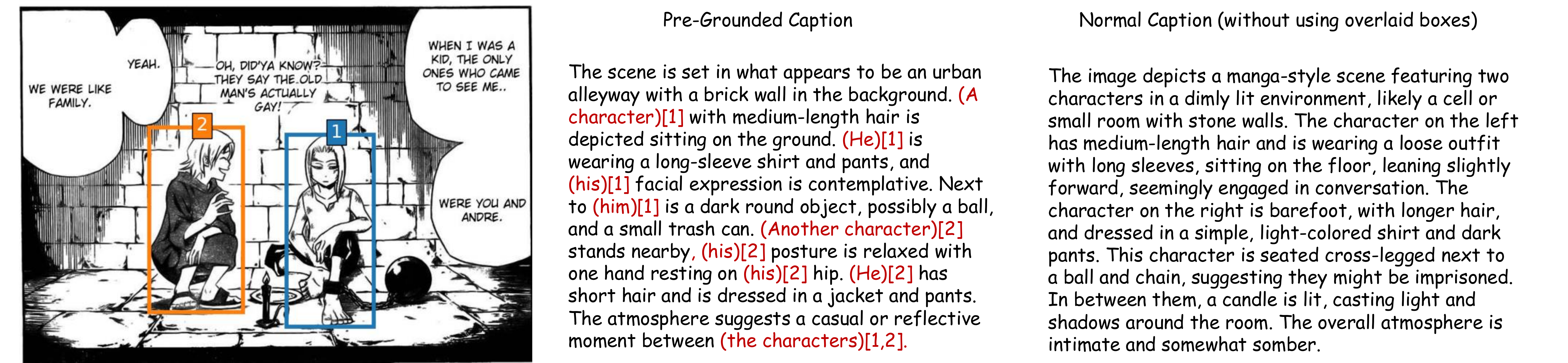}
    \caption{\textbf{Pre-Grounded Captions.} GPT-4o-2024-08-06 is provided with a panel image that has character bounding boxes overlaid. The model is then instructed to caption the image, while outputting any character-referring phrases in \texttt{( phrase ) [ ID ]} format. Note that while the characters have been successfully grounded in the predicted caption (left), the quality of this caption is significantly inferior when compared to the normal caption (right) which is obtained without overlaying the boxes. Image: Belmondo by Ishioka Shouei.}
    \label{fig:zero_shot_character_grounding}
\end{figure*}

In this prompt, we explicitly instruct GPT-4o-2024-08-06 to pre-ground the characters by generating captions that include character-referring phrases in a predefined format that links the overlaid bounding box ID to the character-referring phrase. This ensures the output is structured and ready for automated parsing. An example is illustrated in~\cref{fig:zero_shot_character_grounding}. Once GPT-4o-2024-08-06 generates the captions, we parse the output according to the predefined format, extracting the grounded character annotations. This process allows us to effectively mine character grounding information in a zero-shot manner.

To generate training data for Magiv3, we apply this method to panels from the PopManga dev set. The annotations mined through this pipeline are then used to train Magiv3.

\noindent\textbf{Limitations:} There are a few limitations of this method, which are thus reflected in Magiv3's abilities. First, there is a very obvious and significant loss in caption quality, compared to directly asking GPT-4o-2024-08-06 to caption the raw panel image. This is likely because the interleaved brackets are unnatural for typical captions and negatively condition future tokens. Second, the model struggles in crowded scenes where multiple characters are present in close proximity or overlapping. Overlaying the bounding boxes on such panels biases GPT-4o-2024-08-06's interpretation by making all characters appear equally important, even if many are irrelevant. This often leads the model to describe every character in detail, resulting in verbose and less informative captions. Without the bounding boxes, the model is typically more effective at identifying the central focus of the panel and ignoring minor or background characters. However, the bounding boxes disrupt this natural prioritisation, deteriorating the quality of the captions and their utility for training data generation. Additionally, in densely packed panels with many characters, GPT-4o-2024-08-06 struggles to resolve ambiguities, leading to incorrect IDs matched to referring phrases. Finally, GPT-4o-2024-08-06 occasionally fails to adhere to the predefined output format, resulting in parsing errors and necessitating manual intervention or re-computation. These issues collectively make this approach both noisy and resource-intensive, particularly when scaling to larger datasets.

\section{More on PopCaptions}
As discussed in the main paper, PopCaptions is a novel dataset of 3,374 captioned manga panels, drawn from 15 randomly selected chapters of the PopManga test set~\cite{magiv1} spanning 694 manga pages. Additionally for each of these pages, we provide the overall ground truth story.

The captioning was performed as a two-step annotation process to ensure quality in panel descriptions and character identification. Initially, captions were generated using gpt-4o, offering a solid foundation of scene, character, and emotion descriptions, though these AI-generated captions often contained errors, embellishments, and occasional inconsistencies, especially regarding character actions and emotions.

To refine these captions, a team of four annotators reviewed and edited each one, correcting misinterpretations, removing inaccuracies, and enhancing clarity in character references. Annotators also selected character-referencing phrases and drew bounding boxes around corresponding characters in each panel, linking text descriptions to visual elements. This detailed manual process took approximately 3 minutes per panel, totaling over 10,000 minutes across 3,374 panels, ensuring a high-quality final dataset.

To obtain the page-level stories, we utilised the same pipeline proposed in this paper i.e.\ zero-shot prompting of an LLM, except all the information provided in the prompt was ground truth (annotated by humans) and therefore the LLM was only required to compile this information into a prose format. Specifically, we construct a prompt (detailed in~\cref{sec:prompts} below) using human annotated captions, human annotated transcripts with ground truth character names, and instruct Deepseek-R1~\cite{deepseekr1} to generate the story. We then randomly sampled a subset of these for manual quality assurance and found the generated stories to be of high quality.

\section{More on prompts}
\label{sec:prompts}
In this paper, we employed several prompts tailored to different tasks, such as prose generation (Sec. 5.2 of the main paper) and evaluation using language model judges (Sec. 7.3 of the main paper). These prompts were crucial for guiding the LLMs to produce outputs aligned with our objectives.

We do not claim that our prompts are optimal. The process of prompt design allows for a degree of artistic variation, and the vast space of possible prompts makes exhaustive ablation infeasible. Additionally, evaluating all variations is computationally and financially expensive. Despite this, we document the exact prompts we used to maintain transparency and reproducibility.

\subsection{Prompt for prose generation}

\begin{mdframed}[backgroundcolor=lightgray, linewidth=1pt, roundcorner=10pt]
\scriptsize
\begin{verbatim}
I have a series of manga panel descriptions and dialogues.

Panel 1

Description: {caption}
Dialogues: {transcript}

Panel 2

Description: {caption}
Dialogues: {transcript}

...

Panel N

Description: {caption}
Dialogues: {transcript}

I want you to write a summary so that a blind or visually impaired person can understand the story. 
Make sure to stick to the provided details. All these panels belong to the same page so make sure 
your narrative is coherent. The format of the narrative should be a prose.
\end{verbatim}
\end{mdframed}

\newpage
\subsection{Prompt for LLM judges}
\begin{mdframed}[backgroundcolor=lightgray, linewidth=1pt, roundcorner=10pt]
\scriptsize
\begin{verbatim}
You are trying to tell if a candidate caption/prose is describing the same image as a reference caption/prose. 
Given the following rubric, I want you to give me a score on a scale from 1-5.

### Rubric for Evaluating Manga Panel Descriptions (1-5 Scale)

1. **Severely Inaccurate (1):**
   - The predicted caption/prose is mostly unrelated to the reference. Key elements are either missing or
   incorrectly presented, and there are major contradictions that obscure the intended context.

2. **Somewhat Off-Base (2):**
   - The predicted caption/prose captures some correct ideas but overlooks many crucial details. Major inaccuracies
   exist, such as incorrect character features or setting descriptions. The overall theme may slightly resemble
   the reference but lacks precision.

3. **Partially Accurate (3):**
   - The predicted caption/prose includes several recognizable aspects of the reference but has important 
   inaccuracies. While it conveys the general idea, significant details about multiple 
   features like characters, actions, or settings may be misrepresented.

4. **Mostly Accurate (4):**
   - The predicted caption/prose captures most key elements of the reference accurately. Minor inaccuracies are 
   permissible as long as they do not significantly alter the overall understanding. Additional thematic elements
   or details may be present if they enhance the scene without conflicting with its primary depiction.

5. **Highly Accurate (5):**
   - The predicted caption/prose is nearly identical to the reference in both content and context. It seamlessly
   captures every detail, and any enhancements serve only to enrich the description without deviating from 
   the original meaning.

Predicted caption/prose:

{}

Reference caption/prose:

{}

Instructions:

Output your response in a json with a key "judgement" that contains your analysis of the predicted caption/prose
and a key "score" between 1 and 5 (decimal is allowed) that contains your score.
\end{verbatim}
\end{mdframed}

\section{Can't industry-grade VLM models do it End-to-End?}
An inquisitive reader might wonder: can't GPT-4o, Gemini etc. handle the entire task end-to-end? For instance, could we simply input a manga page and have the model generate a complete description of the story? This is a very reasonable question, and while the answer is “not quite”, the possibility may not be far off in the future.

\begin{figure*}[h]
    \centering
    \includegraphics[width=0.9\linewidth]{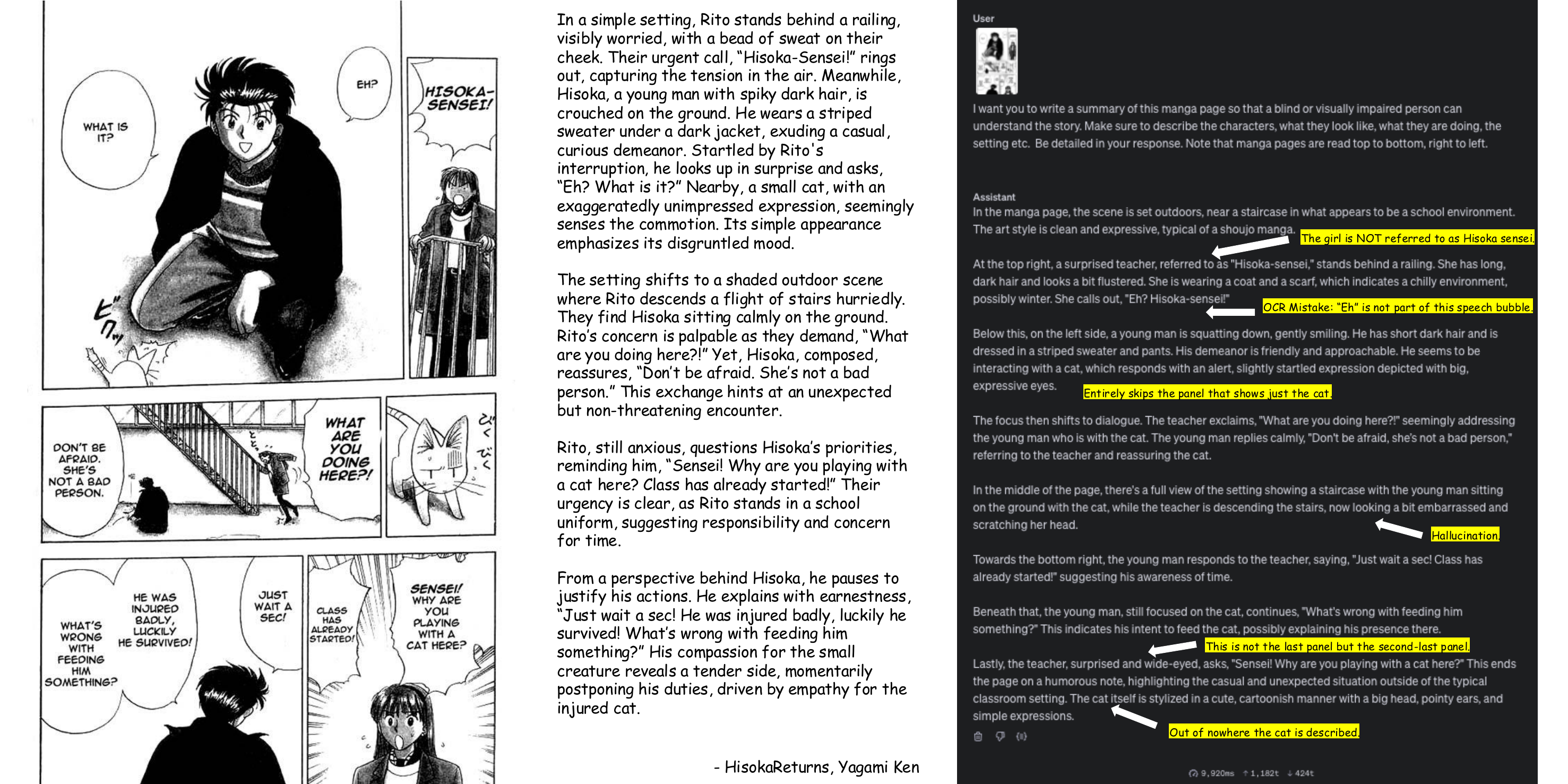}
    \caption{\textbf{Comparison of our pipeline's output (left) vs directly using GPT-4o (right).} Notice the issues highlighted in yellow.}
    \label{fig:end_to_end}
\end{figure*}

\begin{figure*}[h]
    \centering
    \includegraphics[width=0.9\linewidth]{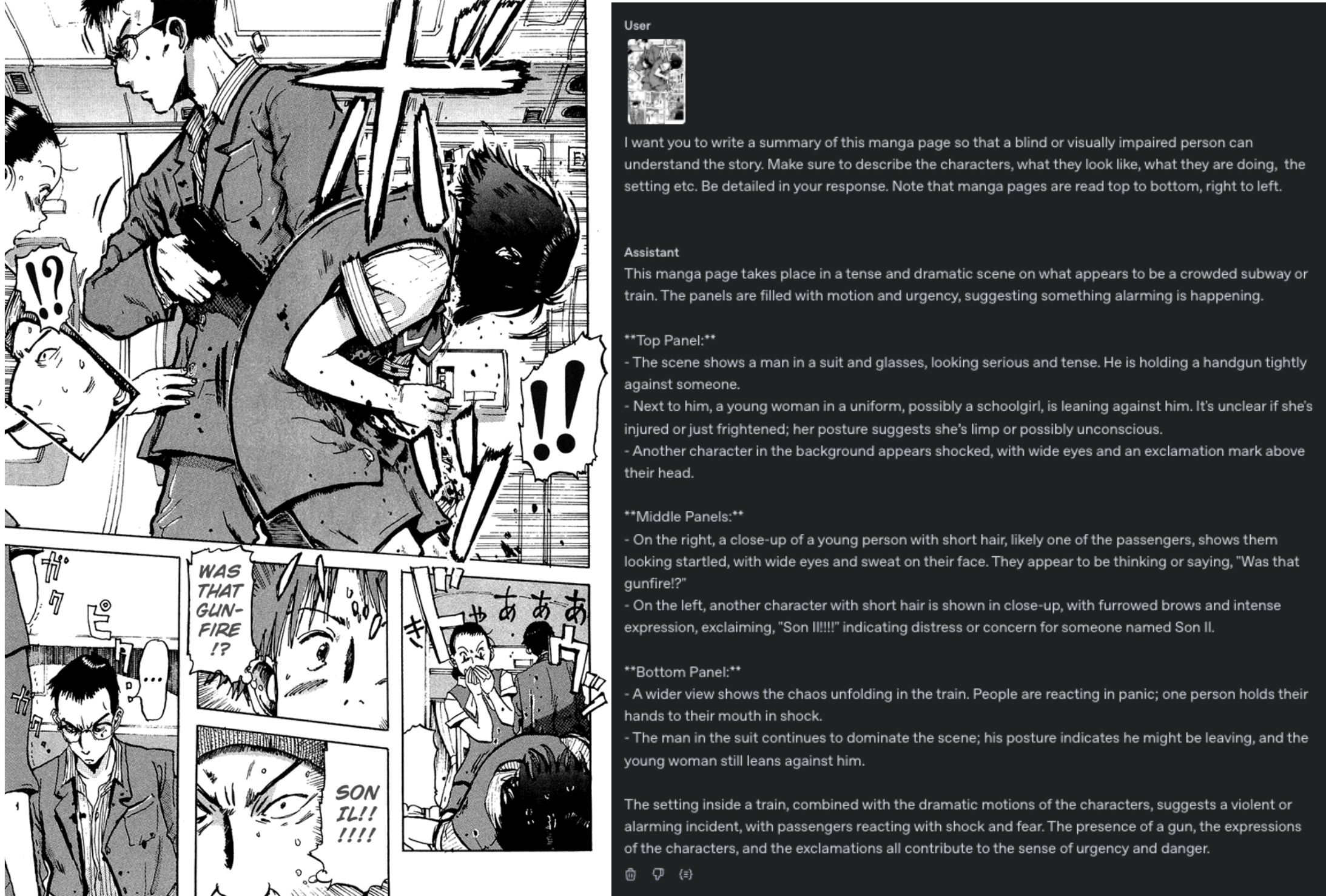}
    \caption{\textbf{Another example of End-to-End.} Ignoring the style of the output, it is apparent that GPT-4o-2024-08-06 is not proficient at clustering characters across panels, and thus struggles with narrative continuity. Additionally, despite the instruction to follow top-bottom, right-to-left, reading order, the model fails to do so. Image: ©HanzaiKousyouninMinegishiEitarou, Kii Takashi.}
    \label{fig:end_to_end_no_clustering}
\end{figure*}

To illustrate, consider an example in~\cref{fig:end_to_end}, where GPT-4o-2024-08-06 is tasked with describing the story directly from a manga page. At first glance, the output may seem satisfactory; however, upon closer inspection, certain obvious issues emerge, as highlighted in the figure. Such challenges highlight the limitations of current models in understanding and conveying the full depth of manga storytelling in a single pass. Furthermore, the chosen example is relatively simple on purpose, featuring minimal action or chaotic elements. Instead, consider the example in~\cref{fig:end_to_end_no_clustering}, where the scene is more complex. We observe that these industrial-grade models do not currently possess the ability to re-identify characters across panels and struggle with narrative continuity.

This is where a step-by-step approach offers distinct advantages. By decomposing the problem, we can incorporate prior knowledge about manga, such as panel reading order or cultural conventions, as well as external resources like a predefined character bank. These inputs allow us to spoon-feed critical details to large models, reducing the likelihood of errors and enabling them to perform at their best. In this sense, we do not view ourselves as competitors to these models but rather as collaborators, integrating them into our pipeline to enhance their output.

Looking ahead, we can envision a future where such models are significantly improved by leveraging data generated from our approach. By using annotations mined through our pipeline, it may become possible to refine these models to better understand complex storytelling formats like manga. Thus, rather than striving for direct replacement, our method serves as a stepping stone toward more capable and reliable AI systems for manga understanding.

\section{Alternative narrative formats}

\begin{figure*}[h]
    \centering
    \includegraphics[width=\linewidth]{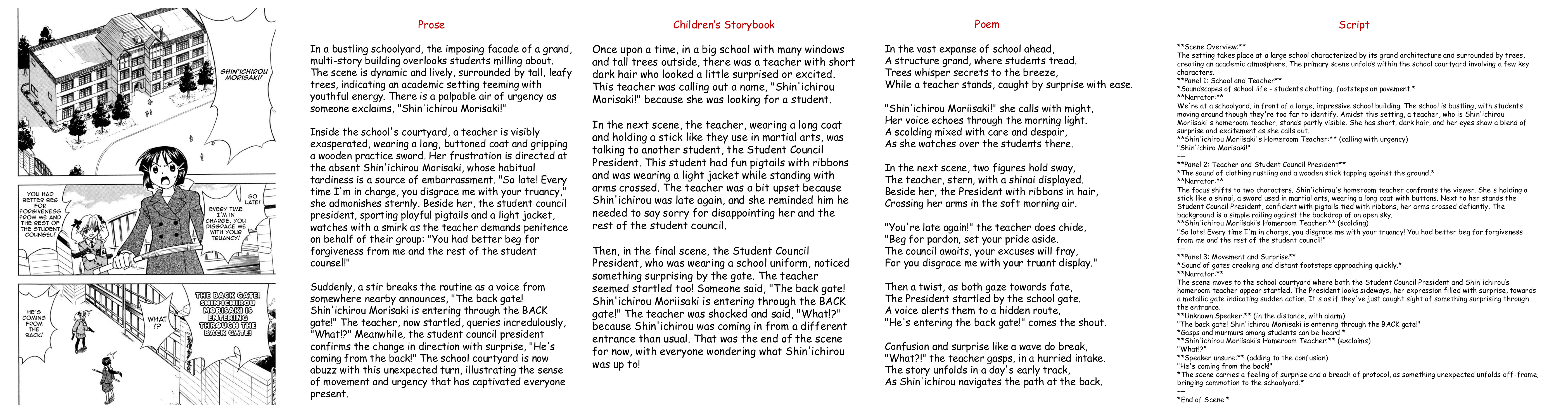}
    \caption{\textbf{Illustration of various possible output formats.} Image: ©Arisa, Yagami Ken.}
    \label{fig:narrative_options}
\end{figure*}

While we chose prose as the primary format to describe manga for visually impaired readers, it is important to note that this choice is arbitrary. The flexibility of leveraging LLMs to generate the final output lies in their ability to adapt to alternative narrative formats, tailored to different audiences or purposes. For instance, as demonstrated in~\cref{fig:narrative_options}, we explore various narrative formats beyond prose. These include a children's storybook format, which simplifies complex manga stories and makes them accessible to younger audiences. Such outputs can also incorporate age-appropriate content moderation, ensuring suitability for children. Additionally, we experiment with generating poetic descriptions, offering a creative and entertaining way to reimagine manga narratives.

Another compelling format is the script, designed to assist animators or film directors in adapting manga into video formats. Scripts can serve as a bridge between static visual storytelling and dynamic visual media, providing structured dialogue, scene descriptions, and action cues.

These examples highlight the versatility of LLMs in generating tailored narrative formats. By aligning the output format with the needs of specific audiences or applications, this approach demonstrates the potential of AI-driven storytelling to go beyond accessibility, offering innovative ways to reimagine manga for diverse contexts.

\section{Copyright and research on comics}
As AI-driven research on comics becomes increasingly prevalent, it is crucial to consider the ethical implications, particularly regarding copyright and fair use. Copyright laws differ across jurisdictions, requiring researchers to carefully assess the legal frameworks governing the use of comic images in their respective countries. In our research, we utilise images scraped from the web, which is legally permitted for non-commercial academic research purposes in our country, aligning with practices used in large-scale datasets such as LAION and KITTI. However, to ensure ethical integrity and respect for creators’ rights, all images included in this manuscript are from Manga109~\cite{manga109} where explicit permission is granted by the authors for use in publications. Researchers in this field should remain vigilant in navigating both legal and ethical considerations to promote responsible AI-driven scholarship on comics.

\section{Examples of LLM scoring}
In this section we provide some examples of reference prose, prose generated using our pipeline, and how the LLM judges critique the predictions.

\newpage
\begin{mdframed}[backgroundcolor=lightgray, linewidth=1pt, roundcorner=10pt]
\tiny
\begin{verbatim}
================================================================================================================================================================
Page: Death Note - c001 - p037 
================================================================================================================================================================
               Reference                |               Predicted                |                                     Scores                                     
================================================================================================================================================================
The scene unfolds with Light Yagami, a  |The page opens with Light Yagami, a     |Chatgpt: 4.5                                                                    
young man with sharp, focused eyes and  |young man with medium-length, textured  |Justification: The predicted prose and the reference prose both describe a      
medium-length hair, standing in tense   |hair, standing pensively. His hand      |scene featuring Light Yagami in deep contemplation, grappling with the          
contemplation. His collared shirt       |rests at his mouth, eyes narrowed in    |revelation of having killed two people. Both descriptions focus on his intense  
sleeve bears a distinct stripe, and a   |deep thought, the lines of his jacket   |expressions, the shadows and silhouettes around him, and his physical actions,  
bead of sweat trails down his cheek as  |sharp against a sparse background.      |such as holding a bag. The predicted prose accurately captures many critical    
he raises a hand to his chin, deep in   |Though his expression is calm, faint    |elements like Light’s emotional turmoil, the bag, and the abstract chaos at     
thought. The plain background           |strokes around his head hint at         |the end. However, there are some discrepancies and additional details in the    
amplifies the gravity of his            |unease. No words break the silence      |predicted prose, such as the detailed description of the bag actions and the    
expression—a mix of concentration and   |here—only the weight of contemplation.  |background’s representation, which differ slightly from the reference. Despite  
unease.    In a close-up, his face      |The scene shifts to his profile,        |these minor differences, the overall depiction and thematic elements align      
tightens, eyes burning with intensity.  |shadows deepening around him. His eyes  |well with the reference.                                                        
The words escape him, raw and           |widen slightly, intensity burning       |                                                                                
fractured: *“I... I’ve killed two       |through their detailed pupils as he     |Gemini: 4.0                                                                     
people...”* His voice hangs in the      |whispers, *"I... I’ve killed two        |Justification: The predicted prose demonstrates a strong understanding of the   
air, trembling with the weight of       |people..."* The admission hangs in the  |reference, capturing most key elements and the overall progression of the       
revelation.    He steps back slightly,  |dark void behind him, stark and         |scene. While there are minor deviations in specific details (e.g., the          
jacket draped over his shoulder, bag    |unresolved.    Suddenly, we see him     |description of Light's clothing, the exact positioning of his hand, the nature  
slipping as he clutches his chest. A    |full-bodied, a bulky bag slung over     |of the abstract shapes in the final panel), these discrepancies do not          
dramatic shadow stretches behind him,   |his shoulder. His hands lift slightly,  |significantly detract from the overall accuracy. The prose effectively conveys  
mirroring his hunched posture like a    |palms upturned in disbelief. A shadow   |the emotional weight and thematic elements of the reference, showcasing         
specter. His speech splinters further:  |on the wall mirrors his pose,           |Light's internal struggle and descent into chaos.  The additions, such as the   
*“I’ve... killed... Two people...       |stretching like a specter. His voice    |"faint strokes" around Light's head and the description of the wall as "cold,   
ME...”* Each phrase fractures, as if    |fractures: *"I’ve... killed..."*—a      |indifferent," enhance the scene without contradicting the original meaning.     
the truth claws its way out against     |pause—*"Two people..."*—then, quieter,  |Therefore, the predicted prose deserves a score of 4, reflecting its high       
his will.    Turning away, his profile  |*"ME..."* The building wall behind him  |level of accuracy and insightful interpretation.                                
reveals sweat glistening on his face,   |feels cold, indifferent to his shock.   |                                                                                
eyes wide with dread. The shadowed      |In the next moment, his face reappears  |Llama: 4.2                                                                      
wall behind him feels oppressive as he  |in profile, sweat beading his temple.   |Justification: The predicted prose is mostly accurate, capturing key elements   
whispers, *“What do I do...”*—a plea    |His collar frames a tense jaw as he     |of the reference such as Light Yagami's emotions, actions, and setting.         
swallowed by the stillness.    His      |mutters, *"What do I do..."* The        |However, there are minor inaccuracies in details such as the description of     
hand suddenly grips a half-open bag,    |background fades to white, isolating    |his hair, jacket, and background.                                               
fingers digging into the fabric. The    |his anxiety.    Action follows: a       |                                                                                
urgency in his voice sharpens:          |close-up of his hands fumbling with a   |Gemma: 4.0                                                                      
*“Get... rid of this evil thing...”*    |zippered bag. Fabric wrinkles under     |Justification: The predicted prose captures the essence of the panel            
The bag’s contents remain hidden, but   |his grip, sleeves rumpled. There’s      |description, hitting all major plot points and character emotions. The          
his desperation is palpable.    The     |urgency here, his fingers trembling as  |language used is evocative and mirrors the tone of the original.  There are     
final moments plunge into chaos.        |he hisses, *"Get... rid of this evil    |some minor variations in detail (e.g., description of Light's pose, the nature  
Abstract shapes swarm the               |thing..."* The bag dominates the        |of the bag's contents), but the overall  narrative arc and mood are faithfully  
page—geometric fragments, jagged        |frame, a focal point of dread.    The   |conveyed.                                                                       
lines, and overlapping patterns         |final panel dissolves into chaos—a      |                                                                                
evoking a fractured mind. In the void,  |jagged blur of shadows and jagged       |                                                                                
Light’s voice cracks, reduced to        |shapes. Screens or machinery loom       |                                                                                
whispers: *“... No...”* The darkness    |abstractly, their forms swallowed by    |                                                                                
swallows him, leaving only the echo of  |high contrast. Light’s voice cuts       |                                                                                
denial.                                 |through the darkness, sharp and final:  |                                                                                
                                        |*"No..."* The page ends suspended in    |                                                                                
                                        |dissonance, his resolve and fear        |                                                                                
                                        |colliding.                              |                                                                                                                                     
\end{verbatim}
\end{mdframed}

\newpage
\begin{mdframed}[backgroundcolor=lightgray, linewidth=1pt, roundcorner=10pt]
\tiny
\begin{verbatim}
================================================================================================================================================================
Page: Dragon Ball - c001 - p025 
================================================================================================================================================================
               Reference                |               Predicted                |                                     Scores                                     
================================================================================================================================================================
The scene opens outdoors where a young  |On a forested outcrop, three figures    |Chatgpt: 4.5                                                                    
boy named Son Goku stands frozen in     |collide in unexpected chaos. A winged   |Justification: The predicted prose closely resembles the reference prose by     
surprise. His spiky hair ruffles        |pterodactyl-like creature, towering     |capturing most of the key elements, including the characters, the pterodactyl-  
slightly as he stares up at a massive,  |and sharp-toothed, coils its spiked     |like creature, and the essence of the scene. Both descriptions successfully     
scaly pterodactyl-like creature         |tail around a young woman named Bulma,  |convey the interaction between Goku, Bulma, and the creature, including the     
looming over him. The beast’s claws     |hoisting her off the ground. Her        |creature's deceptive friendliness, Bulma's distress, and Goku's bemusement      
clutch a terrified young woman, Bulma,  |ponytail swings wildly as she shrieks   |leading to eventual realization. However, some minor discrepancies exist, such  
who screams in panic. The creature      |in alarm, struggling to free herself.   |as the predicted prose adding thematic elements like Bulma's internal           
sneers, its voice booming with mock     |The creature cackles, its voice         |frustration and the forest's atmospheric reactions, which are not explicitly    
politeness: *“Yeah…friend…that’s it!    |booming with false camaraderie:         |mentioned in the reference but enhance the narrative without altering its       
Look, there’s something I have to       |*"Yeah...friend...that's it! Look,      |meaning. Overall, the predicted description maintains the core essence and      
discuss with her in private, okay? You  |there's something I have to discuss     |details of the original scene while introducing additional context that         
just wait right here!”* Goku blinks,    |with her in private, okay? You just     |complements the narrative.                                                      
baffled, as the creature suddenly       |wait right here!"* Below them, a small  |                                                                                
lunges forward, binding him tightly to  |boy with spiky hair—Son Goku—tilts his  |Gemini: 3                                                                       
a nearby tree with ropes.               |head, gripping a staff. His expression  |Justification: The predicted prose follows the general narrative flow of the    
Struggling against the knots, Goku      |flickers between confusion and          |reference but rearranges several key events and misrepresents some details.     
mutters, *“Funny… WHY WOULD HE TIE ME   |curiosity, unanswered questions         |While it captures the core interaction between Goku, Bulma, and the             
UP?”* His confusion lingers as the      |hanging in the air.    The scene        |pterodactyl, the sequence of actions is altered, impacting the portrayal of     
pterodactyl takes flight, its wings     |shifts abruptly. Goku now sits bound    |cause and effect. For example, the predicted prose has Goku tied up after       
thrashing violently. Bulma writhes in   |to a tree trunk, ropes digging into     |Bulma is taken, while the reference clearly shows him being tied up first.      
its grip, her shouts muffled but        |his wrists. Despite his predicament,    |Additionally, some dialogue is misattributed or missing, and the emotional      
desperate: *“WHATA… VHA…”* The          |his face remains eerily calm.           |nuances of the characters are not fully conveyed.  The predicted prose          
creature cackles, *“It’s been far too   |*"Funny..."* he murmurs, more puzzled   |introduces interpretations that are not explicitly present in the reference,    
long since I’ve tasted female human     |than afraid. *"Why would he tie me      |like Goku’s amusement at Bulma's predicament. While these additions don't       
flesh!”* Below, Goku remains tethered,  |up?"* The forest rustles around him,    |necessarily contradict the scene, they add a layer of subjectivity that strays  
his calm demeanor contrasting the       |indifferent to his quiet bewilderment.  |from the original depiction. Although the overall theme is preserved, the       
chaos above.    Bulma’s distress        |Above the treetops, the pterodactyl     |inaccuracies and reordered events prevent the predicted prose from achieving a  
crescendos. Tears stream down her face  |soars, its jagged claws clamped around  |higher level of accuracy. Thus, I am rating this a 3.                           
as she kicks and screams, *“What are    |Bulma’s dress. She dangles              |                                                                                
you WAITING for?!! Rescue me,           |precariously, pigtails whipping in the  |Llama: 4.8                                                                      
idiot!!!!”* Her voice pierces the air,  |wind as the creature’s laughter turns   |Justification: The predicted prose is highly accurate, capturing the key        
urgency radiating from her every word.  |sinister. *"It’s been far too long      |elements of the reference with precision. Minor inaccuracies exist, such as     
Meanwhile, the pterodactyl jeers,       |since I’ve tasted female human          |the order of events, but they do not significantly alter the overall            
*“DON’T TAKE TOO LONG, OKAY?”* as it    |flesh!"* it roars, saliva dripping      |understanding.                                                                  
soars higher.    Goku, still bound,     |from its beak. Bulma’s terrified        |                                                                                
tilts his head thoughtfully. The        |screams (*"WHA... VHA..."*) dissolve    |Gemma: 3.5                                                                      
creature’s taunts—*“imbecile!           |into the rushing air.    Back on the    |Justification: The predicted prose captures the main elements of the            
Dolt!”*—finally click. His eyes widen   |ground, Bulma’s patience snaps. In a    |reference: Goku encountering the pterodactyl, Bulma being kidnapped, and Goku   
with realization. *“Oh, so now she      |burst of frustration, she stomps her    |eventually realizing he needs to rescue her. However, it deviates in some key   
wants me to RESCUE her!”* he exclaims,  |feet, fists clenched and face flushed.  |areas. For example, it embellishes the scene with Goku being bound first, and   
the urgency of Bulma’s cries snapping   |*"What are you WAITING for?!"* she      |Bulma reprimanding Goku for not rescuing her sooner.  The overall plot          
him into focus. The page closes on      |yells into the void, as if Goku could   |progression is similar but has some added details that aren't in the            
this moment of clarity, tension         |hear her across the distance. *"Rescue  |reference.                                                                      
hanging thick between Goku’s            |me, idiot!!!"* Her voice carries a mix  |                                                                                
restrained resolve and the fading       |of fury and desperation.    Goku,       |                                                                                
laughter of the predatory beast.        |finally untangling himself from the     |                                                                                
                                        |ropes, squints skyward. The             |                                                                                
                                        |pterodactyl’s taunting laughter         |                                                                                
                                        |(*"BWA-HA-HA-HA..."*) echoes as it      |                                                                                
                                        |dwindles to a speck. Goku scratches     |                                                                                
                                        |his head, the absurdity dawning on      |                                                                                
                                        |him. *"Oh, so now she wants me to       |                                                                                
                                        |rescue her!"* he mutters, half-amused,  |                                                                                
                                        |half-resigned. The forest hums with     |                                                                                
                                        |anticipation, leaves whispering as the  |                                                                                
                                        |boy grips his staff tighter, ready—or   |                                                                                
                                        |not—to leap into the fray.              |                                                                                        
\end{verbatim}
\end{mdframed}

\newpage
\begin{mdframed}[backgroundcolor=lightgray, linewidth=1pt, roundcorner=10pt]
\tiny
\begin{verbatim}
================================================================================================================================================================
Page: Jujutsu Kaisen - c001 - p038 
================================================================================================================================================================
               Reference                |               Predicted                |                                     Scores                                     
================================================================================================================================================================
In a dim school corridor late at        |The scene opens with Megumi Fushiguro,  |Chatgpt: 3.5                                                                    
night, Megumi Fushiguro pauses to       |a young man with spiky dark hair and a  |Justification: The predicted prose and the reference prose share several key    
catch his breath, his spiky hair damp   |high-collared uniform, exhaling in      |elements but differ in specific details. Both describe a scene involving        
with sweat and his high-collared coat   |quiet relief. His eyes are closed, a    |Megumi Fushiguro, a conversation with a secondary character, and the presence   
rumpled. His eyes close briefly,        |faint sweat droplet on his cheek, as    |of wolf-like creatures identified as shikigami. However, there are notable      
shoulders slackening in weary relief    |if he’s just weathered something        |differences: the secondary figure is identified as Ryomen Sukuna in the         
after exertion. Nearby, Yuji Itadori    |tense. Behind him, the room is sparse,  |predicted prose and Yuji Itadori in the reference, which suggests a mix-up of   
watches him with a mix of curiosity     |with wooden floors and large windows    |characters. Additionally, the room and corridor settings vary, and the actions  
and skepticism, his light spiky hair    |framing Ryomen Sukuna, a figure with    |and responses of characters show some divergence in tone and context. The       
catching faint moonlight filtering      |short, messy hair crouching nearby.     |predicted prose introduces an additional character, Setsuko Sasaki, not         
through windows.    As Megumi strides   |Megumi walks away dismissively,         |present in the reference. These shifts indicate some inconsistencies with       
forward with renewed determination,     |muttering, “Normally I’d be pissed…     |character identities, settings, and thematic undertones, but the general idea   
motion lines ripple around his feet,    |but good job.” Sukuna retorts sharply,  |and tone are still partially captured.                                          
betraying urgency. "Normally I'd be     |“Why’re you all high and mighty?” A     |                                                                                
pissed," he admits gruffly, pausing to  |soft *fwoosh* sound lingers in the      |Gemini: 2.0                                                                     
glance at Yuji, "...but good job."      |air, hinting at unseen activity.        |Justification: While the predicted prose captures the general storyline and     
Yuji frowns, arms crossed. "Why're you  |The focus shifts to a hallway where     |some key dialogue, it suffers from significant character misidentification and  
all high and mighty?" he retorts,       |Megumi stands calmly beside two wolf-   |inaccuracies in the setting and action details.  Megumi's relief after          
though his tone lacks malice.    The    |like creatures—one white, one           |exertion, the presence of Yuji (misidentified as Ryomen Sukuna and later        
tension shifts as two shadowy           |dark—sniffing at something invisible.   |Setsuko Sasaki), and the dynamic between them are core to the scene but are     
creatures materialize: one wolf-like    |Behind him, a light-haired person       |misrepresented. The description of Sukuna in a hooded sweatshirt is also        
with sharp ears devours a writhing      |named Setsuko Sasaki cradles an         |incorrect. Although some thematic elements are preserved, like the discussion   
mass of darkness, while another, paler  |unconscious figure in their arms,       |about curses and shikigami, the overall context is skewed by these errors.      
beast gnashes its teeth nearby. Yuji    |their expression curious. Another       |                                                                                
blinks at the spectacle. "By the way,   |voice asks, “By the way, what’s that    |Llama: 4.2                                                                      
what's that eating the curse?" Megumi   |eating the curse?” Megumi replies,      |Justification: The predicted prose is mostly accurate, capturing the key        
replies flatly, "It's my shikigami.     |“It’s my shikigami. You can see it?”    |elements of the reference prose. However, there are some inaccuracies, such as  
You can see it?" Behind them, a small   |The question hangs unanswered.          |the incorrect character name (Ryomen Sukuna instead of Yuji Itadori) and some   
girl with glasses lies unconscious in   |Megumi turns serious, explaining,       |minor details like the absence of motion lines around Megumi's feet.            
Yuji's arms, her stillness contrasting  |“Normally, you can’t see Curses.        |Additionally, the predicted prose omits some details like the moonlight         
the chaos.    Megumi’s gaze drops, his  |Exceptions being when facing death or   |filtering through the windows and the corridor humming with residual energy.    
voice lowering as if lecturing          |in special places like this.” Sukuna,   |                                                                                
himself. "Normally, you can't see       |now wearing a hooded sweatshirt, looks  |Gemma: 2.0                                                                      
Curses. Exceptions being when facing    |startled, admitting, “Ohh… now that     |Justification: The predicted prose significantly alters the characters and      
death or in special places like this."  |you mention it, I’ve never seen a       |their interactions.  Megumi and Sukuna are swapped for Megumi and Sukuna, and   
Yuji tilts his head, adjusting his      |ghost before.” His gaze drifts to       |Setsuko Sasaki is introduced.  Furthermore, the scene with Yuji and the         
grip on the girl. "Ohh... now that you  |Setsuko, who remains still, glasses     |unconscious girl is absent. While it captures the general mood, the specific    
mention it, I've never seen a ghost     |askew, as if asleep.    The tension     |details and character relationships are inaccurate.                             
before," he muses, unfazed.    A beat   |shifts as Sukuna, leaning against a     |                                                                                
passes. Megumi studies Yuji’s calm      |hallway wall, challenges Megumi:        |                                                                                
demeanor. "You're not scared, huh?"     |“You’re not scared, huh?” Megumi faces  |                                                                                
Yuji’s expression softens, his focus    |him coolly, hands likely tucked in his  |                                                                                
drifting. "Well, I was..." He meets     |dark coat. Sukuna adds, “Well, I was…   |                                                                                
Megumi’s eyes, voice steady. "Did you   |Did you know? People actually die.”     |                                                                                
know? People actually die." Megumi      |Megumi’s brow furrows in a silent,      |                                                                                
stiffens, caught off guard by the       |wary “Huh?”—the page closing on their   |                                                                                
weight of the statement.    Around      |charged standoff, the wolves and        |                                                                                
them, the corridor hums with residual   |unseen forces lingering like shadows.   |                                                                                
energy—the shikigami finish their       |                                        |                                                                                
task, the girl stirs faintly, and the   |                                        |                                                                                
two young men stand suspended between   |                                        |                                                                                
confrontation and camaraderie, their    |                                        |                                                                                
shared reality etched in shadows and    |                                        |                                                                                
unspoken questions.                     |                                        |                                                                               
\end{verbatim}
\end{mdframed}

\newpage
\begin{mdframed}[backgroundcolor=lightgray, linewidth=1pt, roundcorner=10pt]
\tiny
\begin{verbatim}
================================================================================================================================================================
Page: Naruto - c001 - p021 
================================================================================================================================================================
               Reference                |               Predicted                |                                     Scores                                     
================================================================================================================================================================
The night vibrates with primal energy   |A towering nine-tailed fox demon        |Chatgpt: 4.8                                                                    
as a colossal nine-tailed fox demon     |snarls beneath a full moon, its         |Justification: The predicted prose and the reference prose describe the same    
towers over a trembling forest.         |massive form dwarfing the trembling     |manga page scene with impressive detail and emotional intensity. Both           
Kurama’s massive form blots the         |forest below. Sharp teeth gleam as its  |descriptions capture the colossal presence of Kurama, the nine-tailed fox, and  
moonlit sky, jagged teeth bared in a    |nine swirling tails slice through the   |its impact on the environment, such as the trembling forest and the chaos       
snarl that shakes the earth. Trees      |air like whips, radiating primal fury.  |created by its thrashing movements. Key elements such as the desperation of     
splinter beneath its paws, their        |Below this mythical threat, three       |the shinobi, the critical shout about the arrival of Lord Hokage, and the       
trunks snapping like twigs as debris    |shinobi brace themselves on a           |emotional turmoil of Iruka Umino are similarly captured in both versions.       
whirls in chaotic spirals around its    |fractured battlefield. Two crouch low,  |While there are slight variations in wording, the thematic elements and         
thrashing tails. A guttural growl       |their headbands glinting as they grit   |actions described are highly consistent with each other. Additionally, both     
rumbles from its throat—*“Vy…”*—a       |their teeth against an unseen force,    |versions conclude with a shift to focusing on a singular eye, symbolizing       
sound that seems to fracture the air    |while a third is hurled backward mid-   |resolve or tension against the backdrop of destruction. The enhancements and    
itself.    Below, human voices rise in  |air, robes flapping wildly. "We’ve got  |differences present do not obscure the primary depiction of the scene, and      
desperation. A man shouts over the      |to hold on until Lord Hokage gets       |both accounts complement each other well.                                       
din, raw panic straining his words:     |here!" one shouts, voice strained but   |                                                                                
*“WE’VE GOT TO HOLD ON UNTIL LORD       |determined.    Nearby, a young boy      |Gemini: 4.0                                                                     
HOKAGE GETS HERE!”* Bodies              |with spiky hair thrashes against        |Justification: The predicted prose demonstrates a strong understanding of the   
scramble—some lunging forward, others   |invisible restraints, his cloak         |reference image, capturing the central figures, their actions, and the overall  
recoiling—as unseen forces toss a       |billowing as he screams, "Let me go!!"  |atmosphere.  It accurately depicts the nine-tailed fox demon, the struggling    
figure airborne. The ground trembles    |Panic sharpens his features, but        |shinobi, and the distressed boy. The inclusion of dialogue and descriptions of  
with every movement of the beast        |another voice cuts through the chaos—a  |the characters' emotional states enhances the narrative. However, it misses     
above, urgency sharpening the staccato  |companion clinging to him, desperate.   |certain details, such as the specific name of the demon (Kurama), the detailed  
breaths and clattering footsteps of     |"Mom and Dad... they’re still fighting  |destruction caused by the demon, and the name of the boy (Iruka).  The          
those fighting to survive.    Amid the  |the demon!!" The plea hangs in the      |predicted prose also lacks the rich descriptions of sound and motion present    
chaos, Iruka Umino’s voice cracks like  |air, raw and urgent, as motion lines    |in the reference. While it successfully conveys the scene's essence, the        
a whip. He thrashes against arms        |swirl around the boy’s defiant stance.  |omission of these finer points prevents a perfect match.                        
restraining him, his muscles coiled     |The scene shifts abruptly to a single,  |                                                                                
with fury and fear. *“Let me go!!”* he  |piercing eye framed by jagged stones.   |Llama: 4.2                                                                      
demands, the plea tearing from him as   |Its intensity burns with unspoken       |Justification: The predicted prose captures most key elements of the reference  
he strains toward an unseen             |resolve—a silent vow or a moment of     |accurately, but has some minor inaccuracies and creative liberties taken. The   
battleground. *“Mom AND DAD... THEY’RE  |crystallized focus—amid the crumbling   |description of the nine-tailed fox demon, the shinobi, and the young boy's      
STILL FIGHTING THE DEMON!!”* His        |battlefield. The jagged rocks below     |actions are mostly accurate, but some details like the setting, character       
anguish hangs thick, a raw thread       |the eye mirror the fractured world      |features, and actions are slightly misrepresented or reimagined.                
woven into the cacophony of crashing    |around them, shadows deepening as the   |                                                                                
trees and distant roars.    Suddenly,   |fox’s roar echoes across the moonlit    |Gemma: 4.0                                                                      
the perspective narrows—a single eye    |night.                                  |Justification: The predicted prose captures most of the key elements of the     
dominates the darkness, wide and        |                                        |reference: the nine-tailed fox demon, the scene of chaos and desperation, the   
unblinking. Shadows carve deep around   |                                        |presence of Iruka Umino, and the image of a resolute eye. It  adds some         
its edges, the pupil sharp with         |                                        |details like the shinobi and their battle strateges, which can be considered    
resolve or dread. Every line of the     |                                        |additions to enhance the scene, but it slightly alters the focus by giving      
face tenses, as if this silent witness  |                                        |more prominence to the shinobi’s actions.                                       
teeters on the brink of a choice that   |                                        |                                                                                
could unravel the storm around them.    |                                        |                                                                                
The world holds its breath.             |                                        |                                                                             
\end{verbatim}
\end{mdframed}

\end{document}